\documentclass[10pt,journal,compsoc]{IEEEtran}

\ifCLASSOPTIONcompsoc
  \usepackage[nocompress]{cite}
\else
  \usepackage{cite}
\fi

\ifCLASSINFOpdf
\else
\fi

\usepackage{times}
\usepackage{epsfig}
\usepackage{graphicx}
\usepackage{amsmath}
\usepackage{amssymb}
\usepackage{booktabs}
\usepackage{multirow}
\graphicspath{{fig/}}
\usepackage[pagebackref=true,breaklinks=true,colorlinks,bookmarks=false]{hyperref}
\usepackage{float}

\usepackage{tabularx}

\begin{document}
%\title{SnowflakeNet: Point Cloud Completion by Snowflake Point %Deconvolution\\ with Skip-Transformer}
\title{Snowflake Point Deconvolution for Point Cloud Completion and Generation with Skip-Transformer}

\author{Peng~Xiang\IEEEauthorrefmark{1}, Xin~Wen\IEEEauthorrefmark{1}, Yu-Shen~Liu~\IEEEmembership{Member,~IEEE}, Yan-Pei~Cao, Pengfei~Wan, Wen~Zheng, Zhizhong~Han

\IEEEcompsocitemizethanks{
\IEEEcompsocthanksitem Peng Xiang and Xin Wen are with the School of Software, Tsinghua University, Beijing, China. Xin Wen is also with JD.com, Beijing, China. E-mail: xp20@mails.tsinghua.edu.cn, wenxin16@jd.com
%\IEEEcompsocthanksitem Xin Wen is with JD.com, Beijing, China. E-mail: wenxin16@jd.com
\IEEEcompsocthanksitem Yu-Shen Liu is with the School of Software, BNRist, Tsinghua University, Beijing, China. E-mail: liuyushen@tsinghua.edu.cn
\IEEEcompsocthanksitem Yan-Pei Cao is with the ARC Lab, Tencent PCG. E-mail: caoyanpei@gmail.com
\IEEEcompsocthanksitem Pengfei Wan and Wen Zheng are with the Y-tech, Kuaishou Technology, Beijing, China. E-mail: \{wanpengfei, zhengwen\}@kuaishou.com
\IEEEcompsocthanksitem Zhizhong Han is with the Department of Computer Science, Wayne State University, USA. E-mail: h312h@wayne.edu}% <-this % stops an unwanted space
\thanks{Peng Xiang and Xin Wen contributed equally to this work. Yu-Shen Liu is the corresponding author. This work was supported by National Key R\&D Program of China (2022YFC3800600), and the National Natural Science Foundation of China (62272263, 62072268), and in part by Tsinghua-Kuaishou Institute of Future Media Data. Source code is available at \url{https://github.com/AllenXiangX/SnowflakeNet}.}

}

\markboth{Journal of \LaTeX\ Class Files,~Vol.~14, No.~8, August~2015}%
{Shell \MakeLowercase{\textit{et al.}}: Bare Advanced Demo of IEEEtran.cls for IEEE Computer Society Journals}



\IEEEtitleabstractindextext{%
\begin{abstract}
Most existing point cloud completion methods suffer from the discrete nature of point clouds and the unstructured prediction of points in local regions, which makes it difficult to reveal fine local geometric details. To resolve this issue, we propose SnowflakeNet with snowflake point deconvolution (SPD) to generate complete point clouds. SPD models the generation of point clouds as the snowflake-like growth of points, where child points are generated progressively by splitting their parent points after each SPD. Our insight into the detailed geometry is to introduce a skip-transformer in the SPD to learn the point splitting patterns that can best fit the local regions. The skip-transformer leverages attention mechanism to summarize the splitting patterns used in the previous SPD layer to produce the splitting in the current layer. The locally compact and structured point clouds generated by SPD precisely reveal the structural characteristics of the 3D shape in local patches, which enables us to predict highly detailed geometries. Moreover, since SPD is a general operation that is not limited to completion, we explore its applications in other generative tasks, including point cloud auto-encoding, generation, single image reconstruction, and upsampling. Our experimental results outperform state-of-the-art methods under widely used benchmarks. 
\end{abstract}
}

\maketitle

% is not selected <OR> if conference mode is selected - because compsoc
% conference papers position the abstract like regular (non-compsoc)
\IEEEdisplaynontitleabstractindextext

% \fi
\IEEEpeerreviewmaketitle

\IEEEraisesectionheading{\section{Introduction}\label{sec:introduction}}

\begin{figure*}[!t]
    \centering
    \includegraphics[width=1\textwidth]{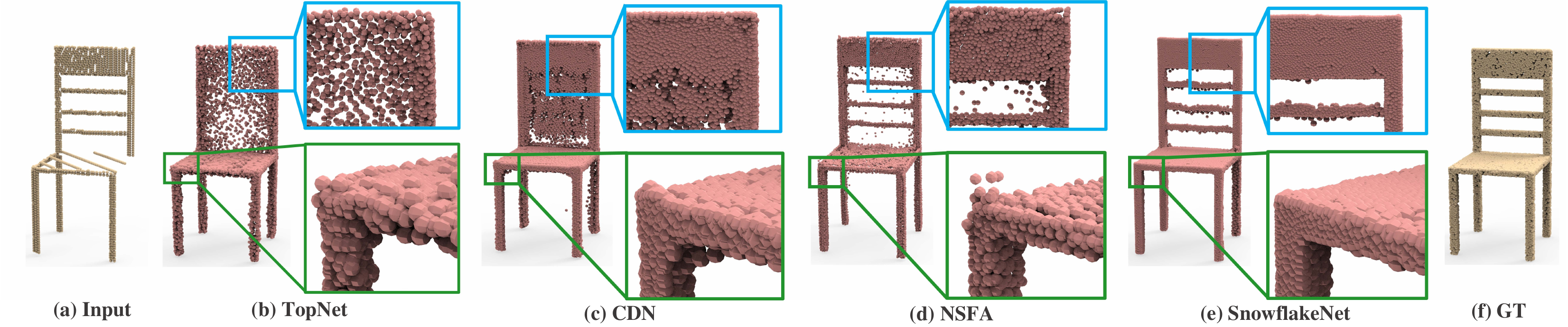}
    \caption{Visual comparison of point cloud completion results. The input and ground truth have 2048 and 16384 points, respectively. Compared with the current completion methods such as TopNet \cite{tchapmi2019topnet}, CDN \cite{wang2020cascaded}, and NSFA \cite{zhang2020detail}, our SnowflakeNet can generate the complete shape (16384 points) with fine-grained geometric details, such as smooth regions (blue boxes), sharp edges, and corners (green boxes).}
    \label{fig:teaser}
\end{figure*}

\IEEEPARstart{I}{n} 3D computer vision \cite{han2020shapecaptioner, han2019view, han20182seq2seq, han2017boscc} applications, raw point clouds captured by 3D scanners and depth cameras are usually sparse and incomplete \cite{wen2020sa,wen2020pmp,wen2020cycle} due to occlusion and limited sensor resolution. Therefore, point cloud completion \cite{wen2020sa,tchapmi2019topnet}, which aims to predict a complete shape from its partial observation, is vital for various downstream tasks. Benefiting from large-scale point cloud datasets, deep learning-based point cloud completion methods have attracted more research interest. Current methods either constrain the generation of point clouds using a hierarchical rooted tree structure \cite{wang2020cascaded,xie2020grnet,tchapmi2019topnet} or assume a specific topology \cite{yang2018foldingnet,wen2020sa} for the target shape. However, most of these methods suffer from the discrete nature of point clouds and the unstructured prediction of points in local regions, which makes it difficult to preserve a well-arranged structure for points in local patches. Therefore, it is still challenging to reveal the local geometric details and structure characteristics, such as smooth regions, sharp edges, and corners, while completing partial 3D shapes, as illustrated in Fig. \ref{fig:teaser}.

\begin{figure}[t]
\begin{center}
   \includegraphics[width=\linewidth]{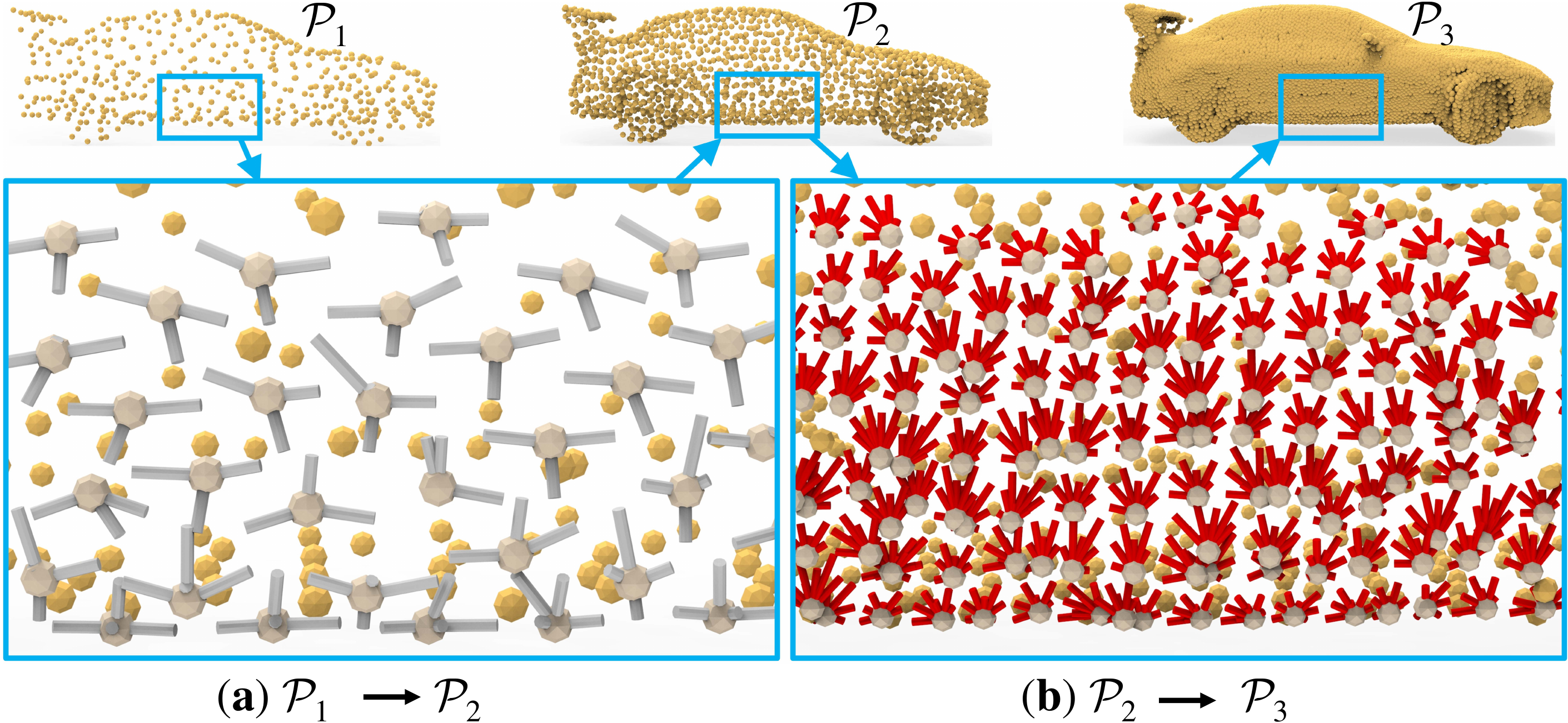}
   \caption{Illustration of snowflake point deconvolution (SPD) for completing missing parts of a car. To show the local changes more clearly, we only illustrate some sample points as parent points in the same patch and demonstrate their splitting paths for child points, which are marked as gray and red lines, respectively. (a) illustrates the SPD of point splitting from a coarse point cloud $\mathcal{P}_1$ (512 points) to its splitting $\mathcal{P}_2$ (2048 points). (b) illustrates the SPD of point splitting from $\mathcal{P}_2$ to complete and dense point cloud $\mathcal{P}_3$ (16384 points), where the child points are expanding like the growing process of snowflakes.}
\label{fig:snowflake_conv}
\end{center}
\vspace{-2em}
\end{figure}

To address this problem, we propose a novel network called \emph{SnowflakeNet} that focuses primarily on the decoding process to complete partial point clouds. SnowflakeNet mainly consists of layers of \emph{snowflake point deconvolution} (SPD), which models the generation of complete point clouds like the snowflake growth of points in 3D space. We progressively generate points by stacking one SPD layer upon another. Each SPD layer produces child points by splitting their parent point while inheriting the shape characteristics captured by the parent point. Fig. \ref{fig:snowflake_conv} illustrates the process of SPD and point-wise splitting. 

Our insight into the detailed geometry is to introduce a \emph{skip-transformer} in SPD to learn the point splitting patterns that can best fit the local regions. Compared with the previous methods, which often ignore the spatial relationship among points \cite{yang2018foldingnet,tchapmi2019topnet,liu2020morphing} or merely learn through self-attention in a single level of multi-step point cloud decoding \cite{wen2020sa,li2018pu,wang2020cascaded}, our proposed skip-transformer integrates the spatial relationships across different decoding levels. Therefore, it can establish a cross-level spatial relationship between points in different decoding steps and refine their locations to produce a more detailed structure. To achieve this, the skip-transformer leverages the attention mechanism to summarize the splitting patterns used in the previous SPD layer, with the aim of producing splitting in the current SPD layer. The skip-transformer can learn the shape context and the spatial relationship between the points in local patches. This enables the network to precisely capture the structural characteristics in local patches and predict a higher quality point cloud for both smooth planes and sharp edges in 3D space. We achieved state-of-the-art completion accuracy under widely used benchmarks. 

Furthermore, SPD is a general operation for point cloud generation that is simple and effective. We take a step forward and explore its performance in more tasks that are closely related to point cloud generation, including point cloud auto-encoding, novel point cloud generation, single image reconstruction, and point cloud upsampling. The generalization ability of SPD is demonstrated by quantitative and qualitative results in our experiments.

Our main contributions can be summarized as follows.

\begin{itemize}
  \item We propose a novel SnowflakeNet for point cloud completion. Compared with previous methods that do not consider local generation patterns, SnowflakeNet can interpret point cloud completion as an explicit and structured local pattern generation that effectively improves the performance of 3D shape completion.

  \item We propose the novel snowflake point deconvolution (SPD) for progressively increasing the number of points. It reformulates the generation of child points from parent points as a snowflake growing process, where the shape characteristics embedded in the parent point features are extracted and inherited by the child points through a \emph{point-wise splitting} operation.
  \item We introduce a novel skip-transformer to learn the splitting patterns in SPD. It learns the shape context and spatial relationship between child points and parent points. This process encourages SPD to produce locally structured and compact point arrangements and captures the structural characteristics of 3D surfaces in local patches.
  \item In addition to point cloud completion, we further generalize SPD to more tasks related to point cloud generation. With a few network arrangements, SPD can be well applied to multiple point cloud generation scenarios. Comprehensive experiments are conducted to verify the effectiveness and generation ability of SPD.
\end{itemize}

\section{Related Work}

%In this section, we review the methods with respect to both point cloud completion and generation. Section \ref{sec:related_work_completion} reviews the related point cloud completion methods. Section \ref{sec:related_work_generation} reviews the methods in different point cloud generation tasks. Section \ref{sec:related_work_skip_transformer} discusses the relation to transformer.

\subsection{Point Cloud Completion}
\label{sec:related_work_completion}
Point cloud completion methods can be roughly divided into two categories. (1) Traditional point cloud completion methods \cite{sung2015data,berger2014state,thanh2016field,wei2019local} usually assume that the 3D shape has a smooth surface or utilize a large-scaled complete shape dataset to infer the missing regions for incomplete shapes. (2) Deep learning-based methods \cite{huang2020pf,wang2020cascaded,chen2019unpaired, gu2020weakly, NEURIPS2020_ba036d22, hutaoaaai2020, hu2019render4completion, Wang_2021_ICCV_VEPCN, alliegro2021denoise, huang2021rfnet, zhang2021ganinversion, gong2021mepcn, Zhou_2021_ICCVpvdiffusion, zhang2021viewguided, Xie_2021_CVPR_stylebased, wang2021cascaded_self, Jiang2019SDFDiffDRcvpr, han2019parts4feature, han20193d2seqviews, han20193dviewgraph, han2018seqviews2seqlabels, han2020seqxy2seqz, han2020reconstructing, wen2020hvpredictor, wen2020cmpd, wen2019adversarial, han2018deep}, however, learn to predict a complete shape based on the prior of the training data. Our method falls into the second category and focuses on the decoding process of point cloud completion. We briefly review deep learning-based methods below.

\noindent\textbf{Point cloud completion by folding-based decoding.} The development of deep learning-based 3D point cloud processing techniques \cite{han2020drwr, wen2020cf, wen2020point2spatialcapsule, liu2019l2g, liu2019point2sequence, liu2021fine, liu2020lrc, han2019multi, cc2021matching, NeuralPull} has boosted the research of point cloud completion. Due to the discrete nature of point cloud data, the generation of high-quality complete shapes is one of the major concerns in point cloud completion research.
One of the pioneering works is FoldingNet \cite{yang2018foldingnet}, which is proposed for point cloud generation rather than completion. 
%, although it was not originally designed for point cloud completion
It includes a two-stage generation process and combines the assumption that the 3D object lies on a 2D-manifold \cite{tchapmi2019topnet}. Following a similar practice, methods such as SA-Net \cite{wen2020sa} further extended this type of generation process into multiple stages by proposing hierarchical folding in the decoder. However, the problem with these folding-based methods \cite{wen2020sa,yang2018foldingnet,li2019pu, zong2021ashf} is that the 3-dimensional code generated by intermediate layer of the network is an implicit representation of the target shape, which 
cannot be interpreted or constrained to help refine the shape in the local region.
On the other hand, TopNet \cite{tchapmi2019topnet} modeled the point cloud generation process as the growth of a rooted tree, where one parent point feature is projected into several child point features in a feature expansion layer. Similar to FoldingNet \cite{yang2018foldingnet}, the intermediate generation processes of TopNet and SA-Net are also implicit, where the shape information is only represented by the point features and cannot be constrained or explicitly explained.

\noindent\textbf{Point cloud completion by coarse-to-fine decoding.} 
Recently, the explicit coarse-to-fine completion framework \cite{xie2020grnet,dai2017shape, Pan_2021_CVPR_VRCNet, xia2021asfm, Yu_2021_ICCV} has received increasing attention, due to its explainable nature and controllable generation process. Typical methods such as PCN \cite{yuan2018pcn} and NSFA \cite{zhang2020detail} adopted the two-stage generation framework, where a coarse and low-resolution point cloud is first generated by the decoder, and then a lifting module is used to increase the density of point clouds. Such methods can achieve better performance since they can impose more constraints on the generation process of point clouds, i.e., coarse one and the dense one. Followers like CDN \cite{wang2020cascaded} and PF-Net \cite{huang2020pf} further extended the number of generation stages and achieved the current state-of-the-art performance. Although intriguing performance has been achieved by the studies along this line, most of these methods still cannot preserve a locally structured point splitting pattern, as illustrated in Fig. \ref{fig:teaser}. The biggest problem is that these methods only focus on the expansion of the point number and the reconstruction of the global shape, while failing to preserve a well-structured generation process for points in local regions. This makes it difficult for these methods to capture the local detailed geometries and structures of 3D shapes.

Compared with the above methods, our SnowflakeNet takes one step further to explore an explicit, explainable and locally structured solution for generating complete point clouds. SnowflakeNet models the progressive generation of point clouds as a hierarchical rooted tree structure like TopNet, while keeping the process explainable and explicit like CDN \cite{wang2020cascaded} and PF-Net \cite{huang2020pf}. Moreover, it excels over the predecessors by arranging the point splitting in local regions in a locally structured pattern, which enables the precise capture of the detailed geometries and structures of 3D shapes.

\subsection{Point Cloud Generation}
\label{sec:related_work_generation}
In addition to point cloud completion, improvements in 3D point cloud representation learning have significantly stimulated the progress of point cloud generation. Different generative tasks for point clouds can be categorized by the forms of inputs, such as point cloud auto-encoding, point cloud upsampling, single image reconstruction, and novel point cloud generation.

The task of point cloud auto-encoding \cite{yang2018foldingnet, groueix2018atlasnet} aims to learn discriminative representations by encoding the input point cloud into a bottleneck latent code and then reconstructing the point cloud itself from the compact code. Auto-encoding is also commonly used to evaluate the performance of generative decoders \cite{luo2021diffusion, Yang_2019_ICCV_pointflow}. FoldingNet \cite{yang2018foldingnet} is one of the pioneering works that attempts to generate point clouds by transforming regular 2D grids into 3D shape surfaces. AtlasNet \cite{groueix2018atlasnet} takes a step further and uses multiple sets of 2D grids to fit different regions of the underlying shape.

Point cloud upsampling \cite{li2018pu, 3PU} also consumes point clouds, but the inputs are often sparse and nonuniformly distributed, so the target is to generate dense point clouds with the points faithfully and uniformly located on the underlying surface. One of the representative works in this area is PU-GAN \cite{li2019pu}, which leverages a generative adversarial network to bridge the gap between sparse point clouds and dense outputs. More recently, PU-GCN \cite{Qian_2021_CVPR_pugcn} proposed a novel graph convolution network to facilitate upsampling by encoding robust local information. Dis-PU \cite{Li_2021_CVPR} proposes a two-step network to disentangle the targets of upsampling and refinement and achieves credible performance.

The task of single image (view) reconstruction takes a single image as input and aims to generate the underlying point cloud with high quality. Earlier works such as PSGN \cite{fan2017psgn} and AtlasNet \cite{groueix2018atlasnet} attempt to generate point clouds from images through convolution and MLP-based architectures. Recently, 3DAttriFlow \cite{3DAttriFlow} proposed assigning disentangled semantic attributes to 3D point clouds by combining an attribute flow pipe and a deformation pipe, which achieves state-of-the-art performance in single image reconstruction.
%Image-based point cloud generation takes image as input, based on the number of images, it can be divided into single image (view) reconstruction and multi-image (multi-view) reconstruction. 

In addition to deterministic data forms such as point clouds and images, the input can also be a randomly sampled latent code from a probabilistic distribution, which leads to novel shape generation \cite{achlioptas2018learning_pcgan, valsesia2018learning_gcngan, shu20193dtreegan}. To generate point clouds with diverse shapes and good quality, many methods have been explored from diverse perspectives. PointFlow \cite{Yang_2019_ICCV_pointflow} leverages continuous normalizing flow to achieve the transformation from the sampled point distribution to the target. ShapeGF \cite{ShapeGF} moves points to the target position by learning gradient fields. More recently, DPM \cite{luo2021diffusion} takes inspiration from non-equilibrium thermodynamics and conducts point cloud generation through a Markov chain \cite{luo2021diffusion}. 

Despite the different forms of inputs, point cloud generation aims to produce high-quality shapes with detailed geometries. Although decent progress has been made in individual tasks, effective generative operations that can be conveniently applied to different tasks are necessary. Therefore, in this paper, we extend snowflake point deconvolution (SPD) to multiple generative tasks, including point cloud auto-encoding, novel point cloud generation, single image reconstruction, and point cloud upsampling.

\subsection{Relation to Transformer} 
\label{sec:related_work_skip_transformer}
The transformer \cite{vaswani2017attention} was initially proposed for encoding sentences in natural language processing, but soon became popular in the research of 2D computer vision (CV) \cite{dosovitskiy2021an,parmar2018image}. Then, the success of transformer-based 2D CV studies drew the attention of 3D point cloud research, where pioneering studies such as Point Transformer \cite{zhao2021pointtransformer}, PCT \cite{Guo_2021}, and Pointformer \cite{Pan_2021_CVPR} have introduced such framework in the encoding process of point clouds to learn the representation. In our work, instead of only utilizing its representation learning ability, we further extend the application of transformer-based structure into the decoding process of point cloud completion, and reveal its ability to generate high-quality 3D shapes through the proposed skip-transformer.

\begin{figure*}
\begin{center}
\includegraphics[width=\textwidth]{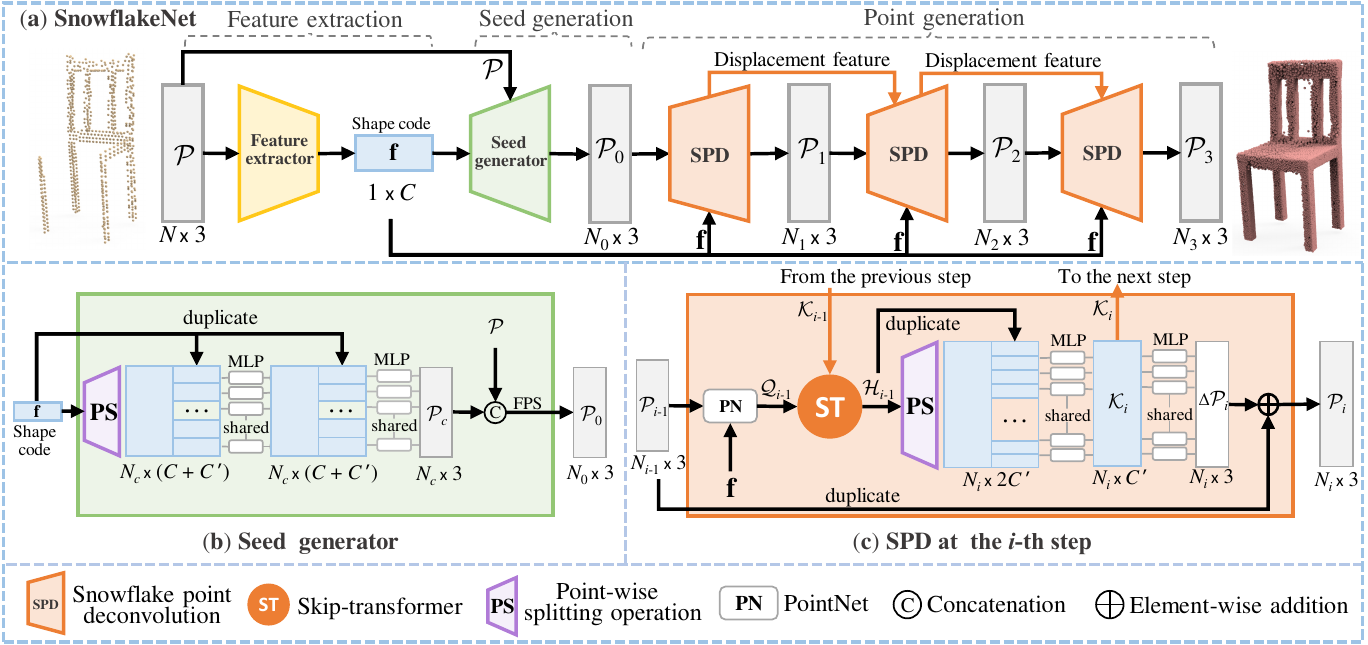}
\end{center}
   \caption{ (a) The overall architecture of SnowflakeNet, which consists of the following three modules: feature extraction, seed generation and point generation. 
   (b) The details of the seed generation module. (c) Snowflake point deconvolution (SPD). Note that $N$, $N_c$, and $N_i$ are the numbers of points, and $C$ and $C'$ are the numbers of point feature channels, which are 512 and 128, respectively.}
\label{fig:overall}
\vspace{-1em}
\end{figure*}

\section{SnowflakeNet}
The overall architecture of SnowflakeNet, which is shown in Fig. \ref{fig:overall} (a), consists of the following three modules: feature extraction, seed generation, and point generation. We will detail each module in the following. 

\subsection{Overview}
\label{sec:method:overview}
\noindent\textbf{Feature extraction module.}
Let $\mathcal{P} = \{ \mathbf{p}_j \}$ with a size of $N \times 3$ be an input point cloud, where $N$ is the number of points and each point $\mathbf{p}_j$ indicates a 3D coordinate. The feature extractor aims to extract a shape code $\mathbf{f}$ with a size of $1 \times C$, which captures the global structure and detailed local pattern of the target shape. To achieve this, we adopt three layers of set abstraction from \cite{qi2017pointnet++} to aggregate point features from local to global, along which point transformer \cite{zhao2021pointtransformer} is applied to incorporate the local shape context.

\noindent\textbf{Seed generation module.}
The objective of the seed generator is to produce a coarse but complete point cloud $\mathcal{P}_0$ with a size of $N_0 \times 3$ that captures the geometry and structure of the target shape. As shown in Fig. \ref{fig:overall} (b), with the extracted shape code $\mathbf{f}$, the seed generator first produces point features that capture both the existing and missing structures through the point-wise splitting operation. Next, the per-point features are integrated with the shape code through a multi-layer perceptron (MLP) to generate a coarse point cloud $\mathcal{P}_c$ of size $N_c \times 3$. Then, following the previous method \cite{wang2020cascaded}, $\mathcal{P}_c$ is merged with the input point cloud $\mathcal{P}$ by concatenation, and then the merged point cloud is downsampled to $\mathcal{P}_0$ through farthest point sampling (FPS) \cite{qi2017pointnet++}.
%Considering that the combined point cloud is unevenly distributed since $\mathcal{P}_c$ and $\mathcal{P}$ overlap with each other in the pre-existing regions, to address this problem, we follow the same tradition of \cite{wang2020cascaded} to 
%down-sampled to $\mathcal{P}_0$ through farthest point sampling (FPS) \cite{qi2017pointnet++}. 
In this paper, we typically set $N_c = 256$ and $N_0=512$, where a sparse point cloud $\mathcal{P}_0$ suffices for representing the underlying shape. $\mathcal{P}_0$ will serve as the seed point cloud for point generation module.

\noindent\textbf{Point generation module.}
The point generation module consists of three snowflake point deconvolution (SPD) steps, each of which takes the point cloud from the previous step as input and splits it by upsampling factors (denoted by $r_1, r_2 $, and $r_3$) to obtain $\mathcal{P}_1, \mathcal{P}_2$, and $ \mathcal{P}_3$, which have the point sizes of $N_1 \times 3, N_2 \times 3$, and $N_3 \times 3$. The SPDs collaborate to generate a rooted tree structure that complies with the local pattern for every seed point. The SPD structure is detailed below.

\begin{figure}
\begin{center}
   \includegraphics[width=0.8\linewidth]{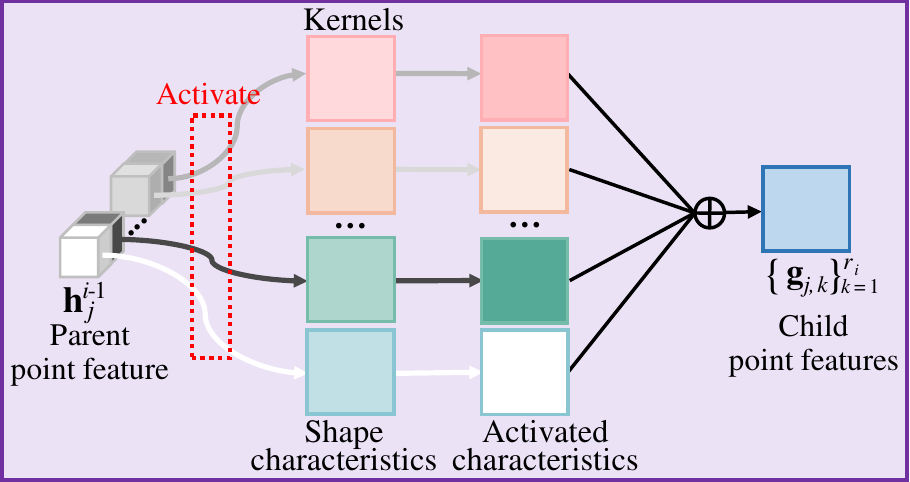}
\end{center}
   \caption{The point-wise splitting operation. The cubes are logits of the parent point feature that represent the activation status of the corresponding shape characteristics (Kernels), and the child point features are obtained by adding activated shape characteristics.}
\label{fig:splitting}
\vspace{-1em}
\end{figure}

\subsection{Snowflake Point Deconvolution (SPD)}
\noindent\textbf{Motivation.} The SPD aims to increase the number of points and reveal the detailed geometry in local regions. This is achieved by multiple steps of point splitting, from which a tree-structured local patch is generated for each seed point. To ensure that the patch can best fit the local region, the following two goals must be achieved in each splitting step: (1) capture the local geometric pattern and (2) transfer the captured geometric information to the child features during splitting. To aggregate accurate local geometric patterns, self-attention is a simple and intuitive solution. In multi-step point splitting, however, self-attention will not suffice because it only considers the context in the current step but ignores the historic splitting information, thus may fail to arrange the multi-step splitting in an organized way. To address this issue, we propose a skip-transformer to enable collaboration between SPDs. While the skip-transformer benefits from the attention mechanism and can well capture the local context, it can also adjust the splitting pattern in consecutive SPDs. After capturing the local geometric pattern, child features can be obtained
%Thesp how to SPD aims to increase the number of points by splitting each parent point into multiple child points, which can be achieved
by first duplicating the parent features and then adding variations. Existing methods \cite{wang2020cascaded,yuan2018pcn,zhang2020detail} usually adopt the folding-based strategy \cite{yang2018foldingnet} to obtain the variations, which is used for learning different displacements for the duplicated points. However, the folding operation samples the same 2D grids for each parent point and ignores the local shape characteristics contained in the parent point. In contrast, the SPD obtains variations through a \textit{point-wise splitting} operation, which fully leverages the geometric information in parent points and adds variations that comply with local patterns. 
%In order to progressively generate the split points, three SPDs are used in point generation module. In addition, to facilitate consecutive SPDs to split points in a coherent manner, we propose a novel skip-transformer to 
%capture the shape context and the spatial relationship between the parent points and their split points.
%enable the SPDs to work in a consistent way.

%\noindent\textbf{Displacements for point splitting.}
Fig. \ref{fig:overall} (c) illustrates the structure of the $i$-th SPD with upsampling factor $r_i$. We denote a set of parent points obtained from the previous step as $\mathcal{P}_{i-1}=\{ \mathbf{p}_j^{i-1}\}_{j=1}^{N_{i-1}}$.
%, where $\mathbf{p}_j^{i-1}$ is the $j$-th parent point in $\mathcal{P}_{i-1}$. 
We split the parent points in $\mathcal{P}_{i-1}$ by duplicating them $r_i$ times to generate a set of child points $\hat{\mathcal{P}}_i$ and then spreading $\hat{\mathcal{P}}_i$ to the neighborhood of the parent points. To achieve this, we take the inspiration from \cite{yuan2018pcn} to predict the \emph{point displacement} $\Delta \mathcal{P}_i$ of $\hat{\mathcal{P}}_i$.
Then, $\hat{\mathcal{P}}_i$ is updated as $\mathcal{P}_i = \hat{\mathcal{P}}_i + \Delta \mathcal{P}_i$, where $\mathcal{P}_i$ is the output of the $i$-th SPD.
%can be generated by $\mathcal{P}_i = \hat{\mathcal{P}}_i + \Delta \mathcal{P}_i$ . 

In detail, taking the shape code $\mathbf{f}$ from feature extraction, the SPD first extracts the per-point feature $\mathcal{Q}_{i-1} = \{\mathbf{q}_j^{i-1}\}_{j=1}^{N_{i-1}}$ for $\mathcal{P}_{i-1}$ by adopting the basic PointNet \cite{qi2017pointnet} framework. Then, $\mathcal{Q}_{i-1}$ is sent to the skip-transformer to learn the \emph{shape context feature}, denoted as $\mathcal{H}_{i-1} = \{\mathbf{h}_j^{i-1}\}_{j=1}^{N_{i-1}}$. 
Next, $\mathcal{H}_{i-1}$ is upsampled by a \emph{point-wise splitting} operation and duplication, respectively, where the former serves to add variations and the latter preserves shape context information. 
%Then, the shape context features are up-sampled by the point-wise splitting operation and duplication to generate a set of features of size $N_i \times 2C'$, where point-wise splitting operation serves to add variations and duplication  serves to preserve shape context information.
%Then, we use point-wise splitting operation and duplication to up-sample $\mathcal{H}_{i-1}$ for the purpose of adding variations to duplicated points and information preserving correspondingly
Finally, the upsampled feature with a size of $N_i \times 2 C'$ is fed to MLP to produce the \emph{displacement feature} $\mathcal{K}_i = \{\mathbf{k}_j^i\}_{j=1}^{N_i}$ of the current step. Here, $\mathcal{K}_i$ is used for generating the point displacement $\Delta \mathcal{P}_i$, and will be fed into the next SPD. $\Delta \mathcal{P}_i$ is formulated as follows:
%Here, $\mathcal{K}_i$ is used for generating the point displacement $\Delta \mathcal{P}_i$, as formulated by
%$\Delta \mathcal{P}_i = \{\Delta \mathbf{p}_j^i\}_{j=1}^{N_i}$
%
\begin{equation}\small
\Delta \mathcal{P}_i = \mathrm{tanh}(\mathrm{MLP}(\mathcal{K}_i)),
\label{eq:displacement}
\end{equation}
where $\mathrm{tanh}$ is the hyper-tangent activation.

\noindent\textbf{Point-wise splitting operation.}
The point-wise splitting operation aims to generate multiple child point features for each $\mathbf{h}_j^{i-1} \in \mathcal{H}_{i-1}$. 
%point-wise splitting operation aims to up-sample the shape context feature $\mathcal{H}_{i-1}$ and add variations
Fig. \ref{fig:splitting} shows this operation structure used in the $i$-th SPD (see Fig. \ref{fig:overall} (c)). It is a special one-dimensional deconvolution strategy, where the kernel size and stride are equal to $r_i$. 
%Such that the operator adopts the same set of kernels to produce child point features for every element in $\mathcal{H}_{i-1}$ in a point-wise manner.
%As shown in Fig. \ref{fig:overall} (c)
In practice, each $\mathbf{h}_j^{i-1} \in \mathcal{H}_{i-1}$ shares the same set of kernels and produces multiple child point features in a point-wise manner.
To be clear, we denote the $m$-th logit of $\mathbf{h}_j^{i-1}$ as $h_{j, m}^{i-1}$, and its corresponding kernel is indicated by $\mathrm{K}_m$. Technically, $\mathrm{K}_m$ is a matrix with a size of $r_i \times \mathrm{C}'$, the $k$-th row of $\mathrm{K}_m$ is denoted as $\mathbf{k}_{m, k}$, and the $k$-th child point feature $\mathbf{g}_{j, k}$ is given by
\begin{equation}\small
\mathbf{g}_{j, k} = \sum_m h_{j, m}^{i-1} \mathbf{k}_{m, k}
\label{eq:dilation}.
\end{equation}
%Then, we generate the child point features $\{\mathbf{g}_{j, k}\}_{k=1}^{r_i}$ of $\mathbf{h}_j^{i-1}$ by
%The point-wise splitting operation first adds variations to duplicated points as follows.
%Given each parent point feature $\mathbf{h}_{j}^{i-1} \in \mathcal{H}_{i-1}$ and convolution kernels $\{\mathrm{K}_m^i\}_{m=1}^D$, 
%where $\mathbf{h}_{j}^{i-1}$ is with $D$ channels and  $\mathrm{K}_m^i$ is the $m$-th kernel that corresponds to the $m$-th channel of $\mathbf{h}_{j}^{i-1}$, the  child point feature $\{\mathbf{g}^{i}_{j, k}\}_{k=1}^{r_i}$ is generated by
%
%\begin{equation}\small
%\{\mathbf{g}_{j, k}\}_{k=1}^{r_i} = \sum_m h_{j, m}^{i-1} \mathrm{K}_m
%\label{eq:dilation},
%\end{equation}
%
%where the sub-script $k$ denotes the $k$-th child.
%where $\mathbf{g}_{j, k}$ denotes the $k$-th child feature of $\mathbf{h}_{j}^{i-1}$, and $\mathbf{h}_{j, (m)}^{i-1}$ denotes the logit of the $m$-th channel. 
In addition, in Fig. \ref{fig:splitting}, we assume that each learnable kernel $\mathrm{K}_m$ indicates a certain shape characteristic, which describes the geometries and structures of the 3D shapes in local regions. Correspondingly, every logit $h_{j, m}^{i-1}$ indicates the activation status of the $m$-th shape characteristic. The child point features can be generated by adding the activated shape characteristics. Moreover, the point-wise splitting operation is flexible for upsampling points. For example, when $r_i = 1$, the SPD can move the point from the previous step to a better position; when $r_i > 1$, it serves to expand the number of points by a factor of $r_i$.

\begin{figure}
\begin{center}
   \includegraphics[width=2.5in]{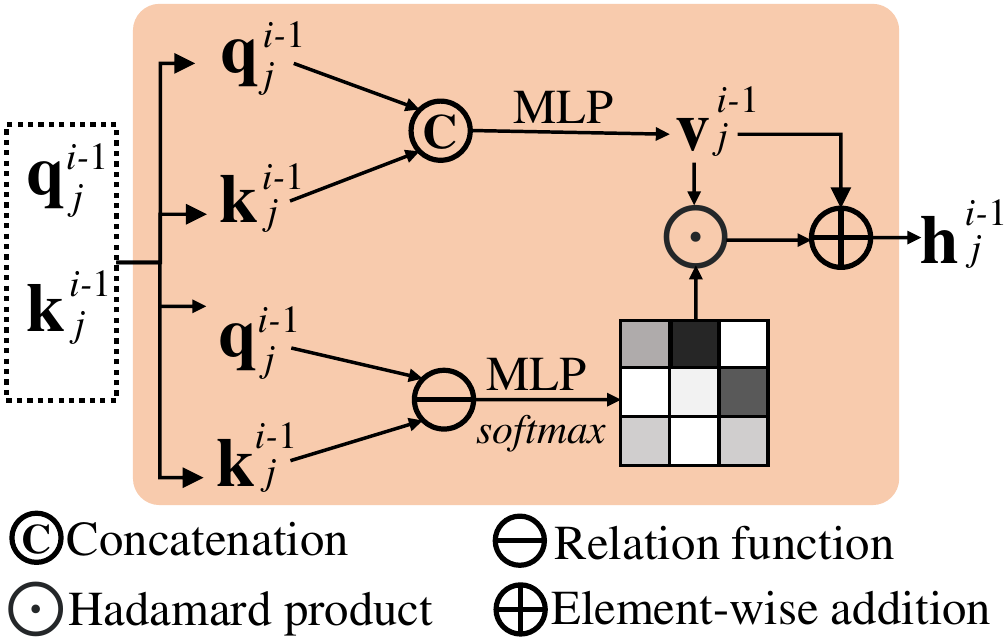}
\end{center}
   \caption{The detailed structure of the skip-transformer.}
\label{fig:skip_transformer}
\vspace{-1em}
\end{figure}

\noindent\textbf{Collaboration between SPDs. }
In Fig. \ref{fig:overall} (a), we adopt three SPDs to generate the complete point cloud. We first set the upsampling factor $r_1=1$ to explicitly rearrange the seed point positions. Then, we set $r_2>1$ and $r_3 > 1$ to generate a structured tree for every point in $\mathcal{P}_1$. Collaboration between SPDs is crucial for coherently growing the tree, because the information from the previous splitting step can be used to guide the current step. In addition, the growth of the rooted trees should also capture the pattern of local patches to keep them from overlapping with each other.
%, otherwise, different trees are very likely to overlap with each other thus degrade quality of the point cloud. 
To achieve this, we propose a novel \emph{skip-transformer} to serve as the cooperation unit between SPDs. In Fig. \ref{fig:skip_transformer}, the skip-transformer takes per-point feature $\mathbf{q}_j^{i-1}$ as input, and combines it with displacement feature $\mathbf{k}_j^{i-1}$ from the previous step to produce the shape context feature $\mathbf{h}_j^{i-1}$, which is given by
\begin{equation}\small
\mathbf{h}_j^{i-1} = \mathrm{ST}(\mathbf{k}_j^{i-1}, \mathbf{q}_j^{i-1}),
\label{eq:skip_transformer}
\end{equation}
%
%displacement feature $\boldsymbol{k}_j^{i-1}$ from previous step as input and combine it with $\boldsymbol{q}_j^{i-1}$ from current step to produce $\boldsymbol{h}_i^{s}$ for the current SPD module, denoted as: 
where $\mathrm{ST}$ denotes the skip-transformer. The detailed structure is described as follows.

\subsection{Skip-Transformer}
\label{sec:skip_transformer}
Fig. \ref{fig:skip_transformer} shows the structure of skip-transformer. The skip-transformer is introduced to learn and refine the spatial context between the parent points and their child points, where the term ``skip'' represents the connection between the displacement feature from the previous layer and the point feature of the current layer.

Given per-point feature $\mathbf{q}_j^{i-1}$ and displacement feature $\mathbf{k}_j^{i-1}$, the skip-transformer first concatenates them. Then, the concatenated feature is fed to MLP, which generates the vector $\mathbf{v}_j^{i-1}$. Here, $\mathbf{v}_j^{i-1}$ serves as the value vector which incorporates previous point splitting information.
%the skip-transformer concatenates and sends them to MLP (note that $\mathbf{q}_j^{i-1}$ and $\mathbf{k}_j^{i-1}$ correspond to the same point $\mathbf{p}_j^{i-1}$). $\mathbf{v}_j^{i-1}$ contains information about previous splitting, 
 To further aggregate local shape context into $\mathbf{v}_j^{i-1}$, the skip-transformer uses $\mathbf{q}_j^{i-1}$ as the query and $\mathbf{k}_j^{i-1}$ as the key to estimate attention vector $\mathbf{a}_j^{i-1}$, where $\mathbf{a}_j^{i-1}$ denotes how much attention the current splitting should pay to the previous one. To enable the skip-transformer to concentrate on local patterns, we calculate attention vectors between each point and its $k$-nearest neighbors ($k$-NN). The $k$-NN strategy also helps to reduce the computational cost. Specifically, given the $j$-th point feature $\mathbf{q}_j^{i-1}$, the attention vector $\mathbf{a}_{j, l}^{i-1}$ between $\mathbf{q}_j^{i-1}$ and the displacement features of the $k$-nearest neighbors $\{\mathbf{k}_{j, l}^{i-1} | l=1,2,\dots,k\}$ can be calculated as follows:
\begin{equation}\small
\mathbf{a}_{j, l}^{i-1} = \frac
{\mathrm{exp}( \mathrm{MLP} ((\mathbf{q}_j^{i-1}) \ominus (\mathbf{k}_{j, l}^{i-1})  )    )}
{\sum_{l=1}^k \mathrm{exp}( \mathrm{MLP} ((\mathbf{q}_j^{i-1}) \ominus (\mathbf{k}_{j, l}^{i-1})  )    )}
\label{eq:attention_vector},
\end{equation}
where $\ominus$ is the relation operation, i.e., elementwise subtraction. Finally, the shape context feature $\mathbf{h}_j^{i-1}$ can be obtained by
\begin{equation}\small
\mathbf{h}_j^{i-1} = \mathbf{v}_{j}^{i-1} \oplus \sum_{l=1}^k \mathbf{a}_{j, l}^{i-1} \odot \mathbf{v}_{j, l}^{i-1}
\label{eq:h},
\end{equation}
where $\oplus$ denotes an elementwise addition and $\odot$ is Hadamard product. Note that there is no previous displacement feature for the first SPD, of which the skip-transformer takes $\mathbf{q}_j^0$ as both the query and key.

\subsection{Training Loss}

\subsubsection{Completion loss}
We leverage the Chamfer distance (CD) or Earth Mover's distance (EMD) as the primary loss function. The $L_2$ version of the Chamfer distance (CD) is defined as follows:
\begin{equation}\small\label{eq:cd}
\mathcal{L}_{\mathrm{CD_2}}(\mathcal{X}, \mathcal{Y}) = \sum_{\mathbf{x} \in \mathcal{X}} \min_{\mathbf{y} \in \mathcal{Y}} \| \mathbf{x}- \mathbf{y} \|_2 + \sum_{\mathbf{y} \in \mathcal{Y}} \min_{\mathbf{x} \in \mathcal{X}} \| \mathbf{y} -\mathbf{x}\|_2,
\end{equation}
where $\mathcal{X}$ and $\mathcal{Y}$ are point sets, $\mathbf{x} \in \mathcal{X}$ and $\mathbf{y} \in \mathcal{Y}$ are point coordinates, respectively. The $L_1$ version of CD replaces $L_2$-norm in Eq. (\ref{eq:cd}) with $L_1$-norm and divide by 2, which is given as 
\begin{equation}\small\label{eq:cd_l1}
\mathcal{L}_{\mathrm{CD_1}}(\mathcal{X}, \mathcal{Y}) = \frac{1}{2}\sum_{\mathbf{x} \in \mathcal{X}} \min_{\mathbf{y} \in \mathcal{Y}} \| \mathbf{x}- \mathbf{y} \| + \frac{1}{2}\sum_{\mathbf{y} \in \mathcal{Y}} \min_{\mathbf{x} \in \mathcal{X}} \| \mathbf{y} -\mathbf{x}\|.
\end{equation}
To explicitly constrain point clouds generated in the seed generation and the subsequent splitting process, we downsample the ground truth point clouds to the same sampling density as $\{\mathcal{P}_c, \mathcal{P}_1, \mathcal{P}_2$, and $\mathcal{P}_3 \}$ (see Fig. \ref{fig:overall}). We define the sum of the four CD losses as the \emph{completion loss}, denoted as follows:
\begin{equation}\small\label{eq:completion_loss}
    \mathcal{L}_{completion} = \mathcal{L}_{\mathrm{CD}} (\mathcal{P}_c , \mathcal{P}_c^{'}) + \sum_{i=1}^3 \mathcal{L}_{\mathrm{CD}} ( \mathcal{P}_i , \mathcal{P}_i^{'} ),
\end{equation}
where $\mathcal{P}_i^{'}$ and $\mathcal{P}_c^{'}$ denote the down-sampled ground truth point clouds that have the same point number as point cloud $\mathcal{P}_i$ and $\mathcal{P}_c$, respectively.

Meanwhile, we also follow MSN \cite{liu2020morphing} and SpareNet \cite{Xie_2021_CVPR_stylebased} to take the Earth Mover's distance (EMD) as training loss, which is defined as follows:

\begin{equation}
    \mathcal{L}_{\mathrm{EMD}} = \min_{\phi:\mathcal{X} \to \mathcal{Y}} \sum_{\mathbf{x} \in \mathcal{X}} \| \mathbf{x} - \phi(\mathbf{x}) \|_2 ,
    \label{eq:emd}
\end{equation}
where $\phi$ is a bijection mapping: $\phi:\mathcal{X} \to \mathcal{Y}$. To take EMD as \emph{completion loss}, we replace the $\mathcal{L}_{\mathrm{CD}}$ in equation \ref{eq:completion_loss} with $\mathcal{L}_\mathrm{EMD}$.

\subsubsection{Preservation loss}
We exploit the \textit{partial matching loss} from \cite{wen2020cycle} to preserve the shape structure of the incomplete point cloud, which is defined as
\begin{equation}\small
\mathcal{L}_{partial}  (\mathcal{X}, \mathcal{Y}) = \sum_{\mathbf{x} \in \mathcal{X}} \min_{\mathbf{y} \in \mathcal{Y}} \| \mathbf{x} - \mathbf{y} \|_2.
\label{eq:partial}
\end{equation}
The partial matching loss is a unidirectional constraint that aims to match one shape to the other without constraining the opposite direction.
Because the partial matching loss only requires the output point cloud to partially match the input, we take it as the \emph{preservation loss} $\mathcal{L}_\mathrm{preservation}$, and the total training loss is formulated as follows:
\begin{equation}\small
\mathcal{L} = \mathcal{L}_\mathrm{completion} + \lambda \mathcal{L}_\mathrm{preservation},
\end{equation}
where we typically set $\lambda=1$.
%
%In our implementation, we use Chamfer distance (CD) as the primary loss function. To explicitly constrain point clouds generated in the seed generation and the subsequent splitting process, we down-sample the ground truth point clouds to the same sampling density as $\{\mathcal{P}_c, \mathcal{P}_1, \mathcal{P}_2, \mathcal{P}_3 \}$ (see Fig. \ref{fig:overall}), where we define the sum of the four CD losses as the \emph{completion loss}, denoted by $\mathcal{L}_\mathrm{completion}$.
%density and obtain the loss of completion $\mathcal{L}_\mathrm{completion}$. 
%Besides, we also exploit the \textit{partial matching loss} from \cite{wen2020cycle} to preserve the shape structure of the input point cloud. It is an unidirectional constraint which aims to match one shape to the other without constraining the opposite direction. Because the partial matching loss only requires the output point cloud to partially match the input, we take it as the \emph{preservation loss} $\mathcal{L}_\mathrm{preservation}$, and the total training loss is formulated as
%

\begin{table}[h]\small
\centering
\tabcolsep=0.1cm
\caption{Point cloud completion on the PCN dataset in terms of per-point $L_1$ Chamfer distance $\times 10^{3}$ (lower is better).}
\resizebox{\linewidth}{!}{\begin{tabular}{l|c|cccccccc}
\toprule
Methods &Avg  &Plane    &Cab.  &Car   &Chair   &Lamp   &Couch    &Table    &Boat   \\ 
\midrule
Folding  \cite{yang2018foldingnet}   &14.31  &9.49    &15.80    &12.61    &15.55   &16.41    &15.97    &13.65    &14.99   \\
TopNet  \cite{tchapmi2019topnet}     &12.15   &7.61   &13.31   &10.90    &13.82    &14.44      &14.78   &11.22  &11.12   \\
AtlasNet \cite{groueix2018atlasnet}   &10.85   &6.37    &11.94    &10.10    &12.06    &12.37    &12.99    &10.33    &10.61   \\
PCN  \cite{yuan2018pcn}   &9.64 &5.50    &22.70    &10.63    &8.70    &11.00    &11.34    &11.68    &8.59   \\
GRNet \cite{xie2020grnet}  &8.83   &6.45   &10.37   &9.45    &9.41    &7.96      &10.51   &8.44  &8.04   \\
CDN \cite{wang2020cascaded}  &8.51   &4.79   &9.97   &8.31    &9.49    &8.94      &10.69   &7.81  &8.05   \\
PMP-Net \cite{wen2020pmp} &8.73  &5.65    &11.24    &9.64    &9.51    &6.95    &10.83    &8.72   &7.25 \\
NSFA \cite{zhang2020detail}  &8.06 &4.76 &10.18 &8.63 &8.53 &7.03 &10.53 &7.35 &7.48 \\
PoinTr\cite{Yu_2021_ICCV}  &8.38 &4.75 &10.47 &8.68 &9.39  &7.75 &10.93 &7.78 &7.29\\
PMP-Net++\cite{wen2022pmp++} & 7.56 & 4.39 & 9.96 & 8.53 & 8.09  &\textbf{6.06} & 9.82 &  7.17 & 6.52 \\
\midrule
Ours	&\textbf{7.21} &\textbf{4.29}	&\textbf{9.16}	&\textbf{8.08}	&\textbf{7.89} &	6.07	&\textbf{9.23}	&\textbf{6.55}	&\textbf{6.40} \\
\bottomrule
\end{tabular}}
\label{table:pcn}
\end{table}

\begin{figure}[h]
\begin{center}
   \includegraphics[width=1.0\linewidth]{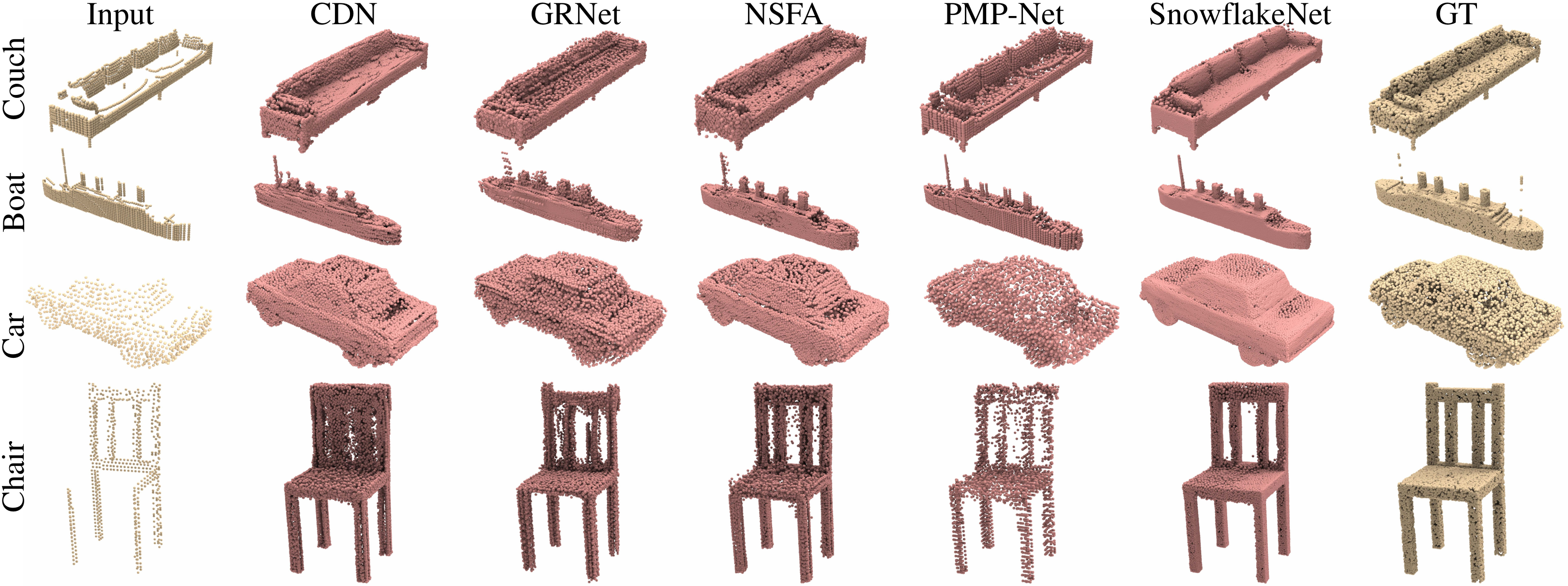}
   \vspace{-1em}
   \caption{Visual comparison of point cloud completion on the PCN dataset. Our SnowflakeNet can produce smoother surfaces (e.g. car) and more detailed structures (e.g. chair back) compared with the other state-of-the-art point cloud completion methods.}
\label{fig:pcn_vis}
\end{center}
\end{figure}

\section{Experiments}
To comprehensively justify the effectiveness of our SnowflakeNet, we first conduct comprehensive experiments under two widely used benchmarks: PCN \cite{yuan2018pcn} and Completion3D \cite{tchapmi2019topnet}. Both are subsets of the ShapeNet dataset. The experiments demonstrate the superiority of our method over other state-of-the-art point cloud completion methods. Moreover, we also conduct experiments on the ShapeNet-34/21 \cite{Yu_2021_ICCV} dataset to evaluate the performance of SnowflakeNet on novel shape completion. Then, we evaluate our method in other point cloud generation tasks, including point cloud auto-encoding, generation, single image reconstruction and upsampling.

\begin{table}[h]\small
\centering
\tabcolsep=0.1cm
\caption{Point cloud completion on the PCN dataset in terms of EMD $\times 10^{2}$ (lower is better).}
\resizebox{\linewidth}{!}{\begin{tabular}{l|c|cccccccc}
\toprule
Methods &Avg  &Plane    &Cab.  &Car   &Chair   &Lamp   &Couch    &Table    &Boat   \\ 
\midrule
Folding \cite{yang2018foldingnet} & 2.526 & 1.682 & 2.576 &  2.183 &  2.847 & 3.062 &  3.003 &  2.500 & 2.357 \\
PCN \cite{yuan2018pcn} & 2.144 & 2.426 &  1.888 & 2.744 & 2.200 & 2.383 & 2.062 & \textbf{1.242} & 2.208 \\
AtlasNet \cite{groueix2018atlasnet} & 2.282 & 1.324 & 2.582 & 2.085 & 2.442 & 2.718 & 2.829 & 2.160 & 2.114 \\
MSN \cite{liu2020morphing} & 2.142 & 1.334 & 2.251 & 2.062 & 2.346 & 2.449 & 2.712 & 1.977 & 2.001 \\
GRNet \cite{xie2020grnet} & 1.987 & 1.376 & 2.128 & 1.918 & 2.127 & 2.150 & 2.468 & 1.852 & 1.876 \\
PMP-Net \cite{wen2020pmp} & 1.863 & 1.259 & 2.058 & 2.520 & 1.798 & \textbf{1.280} & 2.579 & 1.651 & 1.760 \\
SpareNet \cite{Xie_2021_CVPR_stylebased} & 1.862 & 1.131 & 2.014 & 1.783 & 2.050 & 2.063 & 2.333 & 1.729 & 1.790 \\
PointTr \cite{Yu_2021_ICCV} & 1.764 & \textbf{0.938} & 1.986 & 1.851 & 1.892 & 1.740 & 2.242 & 1.931 & 1.532 \\
\midrule
%Ours & \textbf{1.590} & 1.040 & \textbf{1.808} & \textbf{1.706} & \textbf{1.651} & 1.447 & \textbf{2.201} & 1.311 & 1.559 \\
Ours & \textbf{1.532} & 0.973 & \textbf{1.746} & \textbf{1.669} & \textbf{1.614} & 1.419 & \textbf{2.009} & 1.308 & \textbf{1.514} \\

\bottomrule
\end{tabular}}
\label{table:pcn_emd}
\end{table}

\begin{figure}[h]
\begin{center}
   \includegraphics[width=1.0\linewidth]{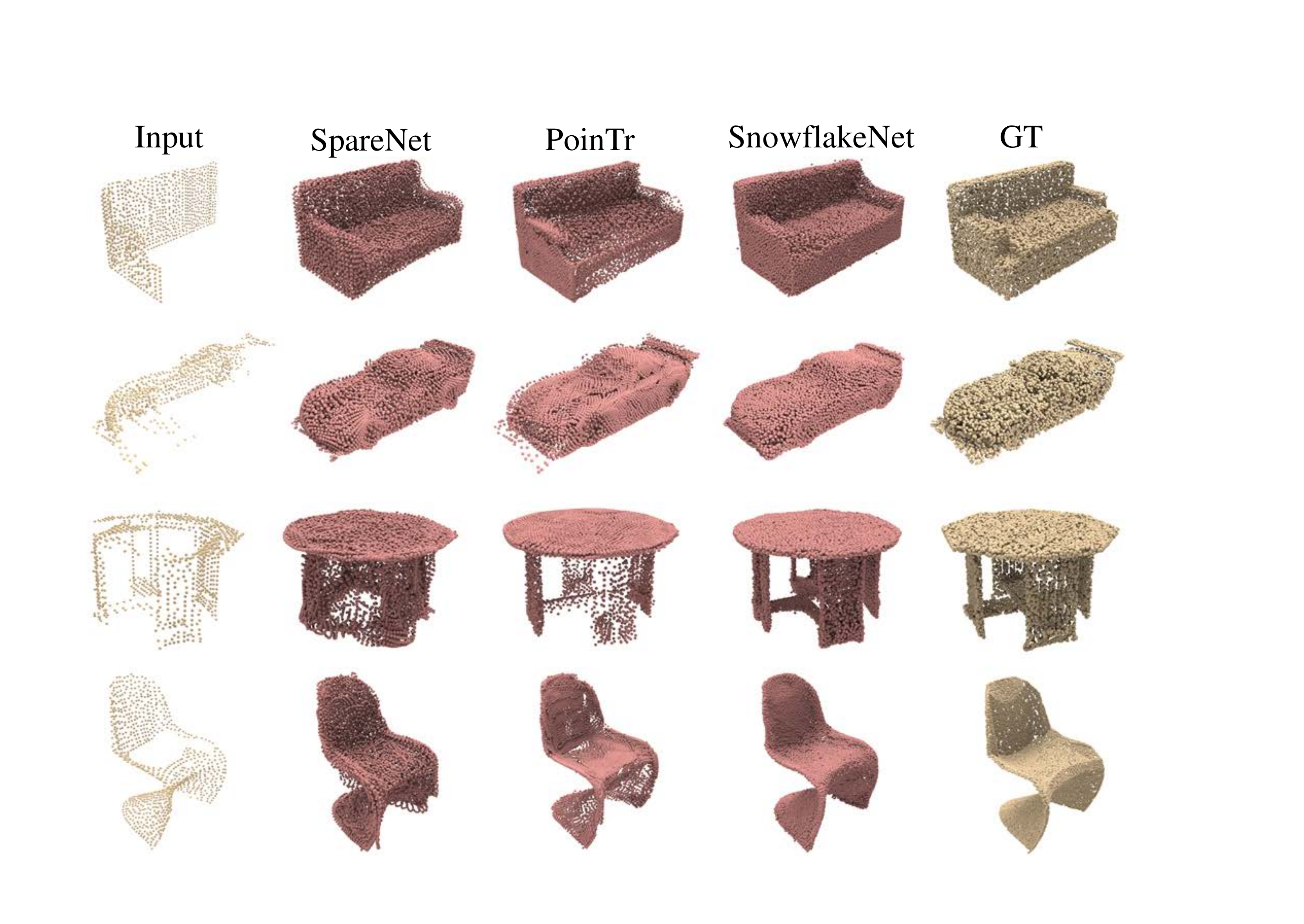}
   \caption{Visual comparison of point cloud completion on the PCN dataset. Following the settings of MSN \cite{liu2020morphing} and SpareNet \cite{Xie_2021_CVPR_stylebased}, SnowflakeNet is trained with EMD.}
\label{fig:vis_pcn_emd}
\end{center}
\end{figure}

\subsection{Evaluation on the PCN Dataset}
\subsubsection{Dataset briefs and evaluation metric}
The PCN dataset \cite{yuan2018pcn} is a subset with eight categories derived from the ShapeNet dataset \cite{chang2015shapenet}. The incomplete shapes are generated by back-projecting complete shapes into eight different partial views. For each complete shape, 16384 points are evenly sampled from the shape surface. To align with the ground truth shapes, we typically set $r_1 = 1$, $r_2 = 4$ and $r_3 = 8$ to generate complete point clouds with 16384 points. We follow the same split settings as PCN \cite{yuan2018pcn} to fairly compare our SnowflakeNet with other methods. For evaluation, we adopt the $L_1$ version of Chamfer distance (CD), which follows the same practice as previous methods \cite{yuan2018pcn}. Although CD is widely used for evaluating the quality of point clouds, it may be insensitive to the global distribution. Therefore, we also use EMD as both the \emph{completion loss} and the evaluation metric to verify the effectiveness of our method.

% \begin{figure*}[h!]
% \begin{center}
%    \includegraphics[width=1.0\textwidth]{fig/pcn_more_pcd.pdf}
%    \caption{Visualization of more completion results using our SnowflakeNet on PCN dataset.}
% \label{fig:pcn_more_vis}
% \end{center}
% \end{figure*}

\subsubsection{Quantitative comparison} Table \ref{table:pcn} shows the results of our SnowflakeNet and other completion methods on the PCN dataset in terms of $L_1$ CD, from which we can find that SnowflakeNet achieves the best performance over all counterparts. In particular, compared with the result of the state-of-the-art PMP-Net++ \cite{wen2022pmp++}, SnowflakeNet reduces the average CD by 0.35, which is 4.6\% lower than the PMP-Net++'s results (7.56 in terms of average CD). Moreover, SnowflakeNet also achieves the best results on most categories in terms of CD, which proves the robust generalization ability of SnowflakeNet
for completing shapes across different categories. In Table \ref{table:pcn}, both CDN \cite{wang2020cascaded} and NSFA \cite{zhang2020detail} are typical point cloud completion methods that adopt a coarse-to-fine shape decoding strategy and model the generation of points as a hierarchical rooted tree. Compared with these two methods, our SnowflakeNet also adopts the same decoding strategy but achieves much better results on the PCN dataset. Therefore, the improvements should be credited to the proposed SPD layers and the skip-transformer in SnowflakeNet, which helps to generate points in local regions with a locally structured pattern.

Table \ref{table:pcn_emd} shows the quantitative results on the PCN dataset in terms of EMD, where SnowflakeNet also achieves the best overall performance (average EMD). In particular, by comparing with the second-ranked PoinTr \cite{Yu_2021_ICCV}, SnowflakeNet is able to reduce the average EMD by 0.232, which is 13.2\% lower than PoinTr (1.764 in terms of average EMD). Compared with CD, the EMD is better at evaluating the global distribution of the completed point clouds \cite{Density_aware}, and the results in Table \ref{table:pcn} and Table \ref{table:pcn_emd} can demonstrate the effectiveness of SnowflakeNet under different evaluation metrics.

\subsubsection{Visual comparison}
We choose the top four point cloud completion methods from Table \ref{table:pcn}, and visually compare SnowflakeNet with these methods in Fig. \ref{fig:pcn_vis}. The visual results show that SnowflakeNet can predict the complete point clouds with a much better shape quality. For example, in the car category, the point distribution on the car's boundary generated by SnowflakeNet is smoother and more uniform than those generated by other methods. For the chair category, SnowflakeNet can predict a more detailed and clear structure of the chair back compared with the other methods, where CDN \cite{wang2020cascaded} almost fails to preserve the basic structure of the chair back, while the other methods generate much noise between the columns of the chair back. 
%In Fig. \ref{fig:pcn_more_vis}, we additionally present more visual completion results on the PCN dataset, which further demonstrates the completion quality of our method. Especially, SnowflakeNet is able to reveal complex local structures like the back cushion of the couch and table rungs.

We also present a visual comparison under the EMD evaluation (see Table \ref{table:pcn_emd}) in Fig. \ref{fig:vis_pcn_emd}. Typically, we visually compare SnowflakeNet with two state-of-the-art methods, SpareNet \cite{Xie_2021_CVPR_stylebased} and PoinTr \cite{Yu_2021_ICCV}. The completion results show that under the guidance of EMD loss, SnowflakeNet can produce point clouds with better visual quality that tend to maintain a consistent point distribution across the entire shape. 

\begin{table}[h]\small
\centering
\tabcolsep=0.1cm
\caption{Point cloud completion on the Completion3D in terms of per-point $L_2$ Chamfer distance $\times 10^{4}$ (lower is better).}
\resizebox{\linewidth}{!}{\begin{tabular}{l|c|cccccccc}
\toprule
Methods &Avg  &Plane    &Cab.  &Car   &Chair   &Lamp   &Couch    &Table    &Boat   \\ 
\midrule
FoldingNet  \cite{yang2018foldingnet}   &19.07  &12.83    &23.01    &14.88    &25.69    &21.79    &21.31    &20.71    &11.51   \\
PCN  \cite{yuan2018pcn}   &18.22 &9.79    &22.70    &12.43    &25.14    &22.72    &20.26    &20.27    &11.73   \\
PointSetVoting \cite{pointsetvoting}  &18.18 &6.88    &21.18    &15.78    &22.54    &18.78    &28.39    &19.96    &11.16   \\
AtlasNet \cite{groueix2018atlasnet}   &17.77   &10.36    &23.40    &13.40    &24.16    &20.24    &20.82    &17.52    &11.62   \\
SoftPoolNet \cite{wang2020softpoolnet} &16.15   &5.81   &24.53   &11.35    &23.63    &18.54      &20.34   &16.89  &7.14   \\
TopNet  \cite{tchapmi2019topnet}     &14.25   &7.32   &18.77   &12.88    &19.82    &14.60      &16.29   &14.89  &8.82   \\
SA-Net \cite{wen2020sa}  &11.22   &5.27   &14.45   &7.78    &13.67    &13.53      &14.22   &11.75  &8.84   \\
%GRNet(2k) \cite{xie2020grnet}  &11.22   &5.7   &16.51   &8.63    &14.48    &11.36      &15.50   &10.67  &6.19   \\
GRNet \cite{xie2020grnet}  &10.64   &6.13   &16.90   &8.27    &12.23    &10.22      &14.93   &10.08  &5.86   \\
PMP-Net \cite{wen2020pmp} &9.23  &3.99    &14.70    &8.55    &10.21    &9.27 &12.43    &8.51   &5.77 \\ 
VRCNet\cite{Pan_2021_CVPR_VRCNet} &8.12 &3.94 &\textbf{10.93} &\textbf{6.44} &9.32 &8.32 &11.35 &8.60 &5.78\\
VE-PCN\cite{Wang_2021_ICCV_VEPCN} &8.10 &3.83 &12.74 &7.86 &\textbf{8.66} &\textbf{7.24} &11.47 &7.88 &\textbf{4.75}\\
\midrule
%PCN-Encoder (Ours) &8.0 &3.82	&12.2	&7.05	&9.21	&\textbf{7.78}	&11.28	&7.12	&\textbf{5.27}	\\
%No PT in Encoder &7.76 &\textbf{3.45}	&11.4	&7.01	&\textbf{8.66}	&8.93	&10.17	&6.68	&5.3 \\
Ours &\textbf{7.60} &\textbf{3.48}	&11.09	&6.90	&8.75	&8.42	&\textbf{10.15}	&\textbf{6.46}	&5.32	\\
\bottomrule
\end{tabular}}
\label{table:completion3d}
\end{table}

\begin{figure}[h]
\begin{center}
   \includegraphics[width=1.0\linewidth]{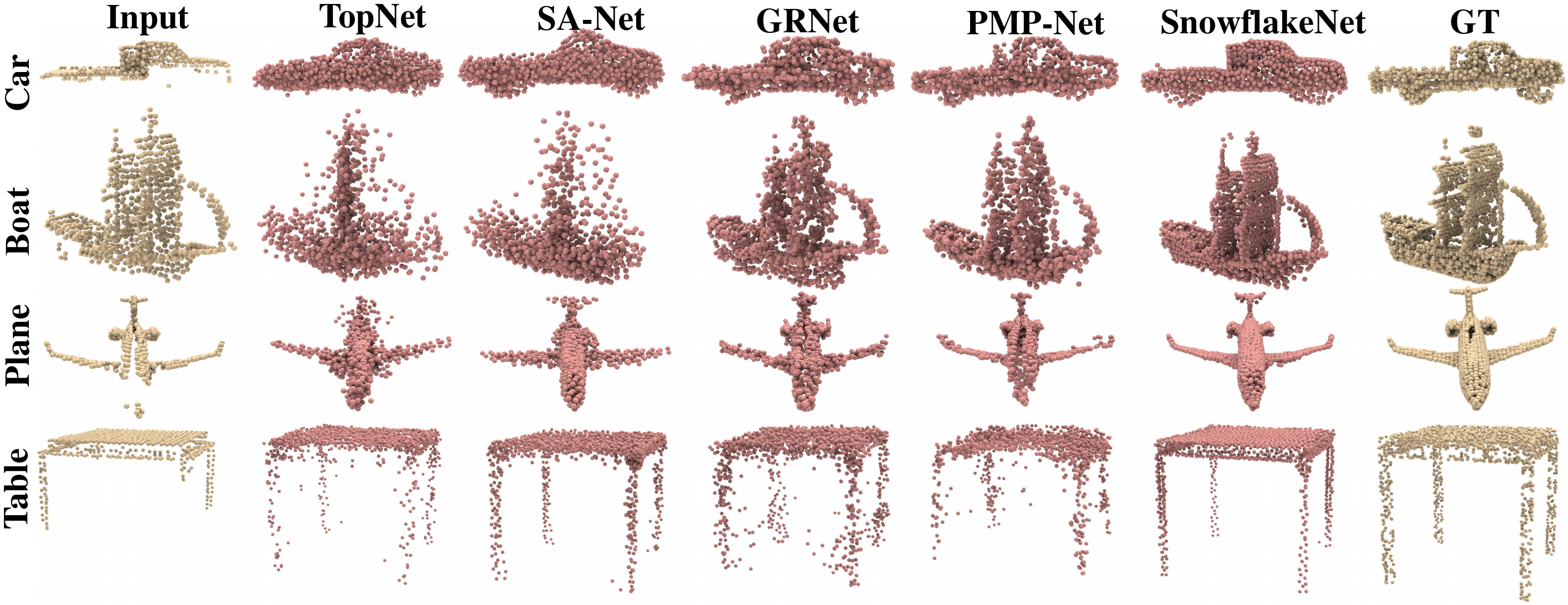}
   \caption{Visual comparison of point cloud completion on the Completion3D dataset. Our SnowflakeNet can produce smoother surfaces (e.g. car and table) and more detailed structures compared with the other state-of-the-art point cloud completion methods.}
\label{fig:completion3d_vis}
\end{center}
\vspace{-2em}
\end{figure}

\subsection{Evaluation on the Completion3D Dataset}\label{sec:compleiton3d}
\subsubsection{Dataset briefs and evaluation metric}
The Completion3D dataset contains 30958 models from 8 categories, of which both the partial and ground truth point clouds have 2048 points, here we set $r_1 = 1$, $r_2 = 2$ and $r_3 = 2$ to generate complete point clouds with 2048 points. We follow the same train/validate/testing split of Completion3D to have a fair comparison with the other methods, where the training set contains 28974 models, the validation and testing sets contain 800 and 1184 models, respectively. For evaluation, we adopt the $L_2$ version of the Chamfer distance on testing set to align with previous studies.

% \begin{figure*}[t!]
% \begin{center}
%    \includegraphics[width=1.0\textwidth]{fig/c3d_more_pcd.pdf}
%    \caption{Visualization of more completion results using our SnowflakeNet on Completion3D dataset.}
% \label{fig:c3d_more_vis}
% \end{center}
% \end{figure*}

\begin{table*}[h]\small
\tabcolsep=0.15cm
\centering
\caption{Point cloud completion on the ShapeNet-34/21 dataset in terms of the $L_2$ Chamfer distance $\times 10^4$ and F-Score@1\% metric. The CD-S, CD-M, and CD-H denote the CD results under the three difficulty levels of \emph{Simple, Moderate, and Hard}.}
\begin{tabular}{l|ccccc|ccccc}
\toprule
 & \multicolumn{5}{c|}{\textbf{34 seen categories}} & \multicolumn{5}{c}{\textbf{21 unseen categories}} \\
 \cmidrule{2-11}
  & CD-S ($\downarrow$) & CD-M ($\downarrow$) & CD-H ($\downarrow$) & CD-Avg ($\downarrow$) & F1  ($\uparrow$) & CD-S ($\downarrow$) & CD-M ($\downarrow$) & CD-H ($\downarrow$) & CD-Avg ($\downarrow$) & F1 ($\uparrow$) \\
  \midrule
  Folding \cite{yang2018foldingnet} & 18.6 & 18.1 & 33.8 & 23.5 & 0.139 & 27.6 & 27.4 & 53.6 & 36.2 & 0.095 \\
  PCN \cite{yuan2018pcn} & 18.7 & 18.1 & 29.7 & 22.2 & 0.154 & 31.7 & 30.8 & 52.9 & 38.5 & 0.101 \\
  TopNet \cite{tchapmi2019topnet} & 17.7 & 16.1 & 35.4 & 23.1 & 0.171 & 26.2 & 24.3 & 54.4 & 35.0 & 0.121 \\
  PFNet \cite{huang2020pf} & 31.6 & 31.9 & 77.1 & 46.8 & 0.347 & 52.9 & 58.7 & 133.3 & 81.6 & 0.322 \\
  GRNet \cite{xie2020grnet} & 12.6 & 13.9 & 25.7 & 17.4 & 0.251 & 18.5 & 22.5 & 48.7 & 29.9 & 0.216 \\
  PoinTr \cite{Yu_2021_ICCV} & 7.6 & 10.5 & 18.8 & 12.3 & \textbf{0.421} & 10.4 & 16.7 & 34.4 & 20.5 & \textbf{0.384} \\
  \midrule
  Ours & \textbf{5.1} & \textbf{7.1} & \textbf{12.1} & \textbf{8.1} & 0.414 & \textbf{7.6} & \textbf{12.3} & \textbf{25.5} & \textbf{15.1} & 0.372 \\
\bottomrule
\end{tabular}
\label{table:shapenet34}
\vspace{-1em}
\end{table*}

\subsubsection{Quantitative comparison}
In Table \ref{table:completion3d}, we show the quantitative results of our SnowflakeNet and those of the other methods on the Completion3D dataset. All results are cited from the online public leaderboard of Completion3D\footnote{https://completion3d.stanford.edu/results}. From Table \ref{table:completion3d}, we can find that our SnowflakeNet achieves the best results over all methods listed on the leaderboard. In particular, compared with the latest point cloud completion methods, such as VE-PCN \cite{Wang_2021_ICCV_VEPCN}, SnowflakeNet can reduce the average CD by 0.50, which is 6.1\% lower than that of the VE-PCN (8.10 in terms of average CD). Meanwhile, SnowflakeNet also outperforms the other methods in multiple categories in terms of per-category CD. Especially in the table category, SnowflakeNet reduces the per-category CD by 1.42 compared with the second-ranked result of VE-PCN.
%compared with the state-of-the-art method PMP-Net \cite{wen2020pmp}, SnowflakeNet significatly reduces the average CD by 1.63, which is 17.3\% lower than the PMP-Net (9.23 in terms of average CD). 
%On the Completion3D dataset, SnowflakeNet outperforms the other methods in all categories in terms of per-category CD. Especially in the cabinet category, SnowflakeNet reduces the per-category CD by 3.61 compared with the second-ranked result of PMP-Net. 
Compared with the PCN dataset, the point clouds in the Completion3D dataset have fewer points, which are easier to generate. Therefore, a coarse-to-fine decoding strategy may have fewer advantages over the other methods. Despite this, our SnowflakeNet achieves superior performance over the folding-based methods such as SA-Net \cite{wen2020sa}, and our method is also the best among the coarse-to-fine methods including TopNet \cite{tchapmi2019topnet} and GRNet \cite{xie2020grnet}. Overall, the results on the Completion3D dataset demonstrate the capability of SnowflakeNet to predict high-quality complete shapes using sparse point clouds.

\subsubsection{Visual comparison}
Similar to the practice in the PCN dataset, we also visually compare SnowflakeNet with the top four methods in Table \ref{table:completion3d}. The visual comparison in Fig. \ref{fig:completion3d_vis} demonstrates that our SnowflakeNet also achieves much better visual results than the other counterparts on a sparse point cloud completion task. Especially, in the plane category, SnowflakeNet predicts the complete plane which is almost the same as the ground truth, while the other methods fail to reveal the complete plane in detail. The same conclusion can also be drawn from the observation of the car category. In the table and boat categories, SnowflakeNet produces more detailed structures compared with the other methods, e.g. the sails of the boat and the legs of the table.

\begin{figure}[h]
\begin{center}
    \centering
    \includegraphics[width=\linewidth]{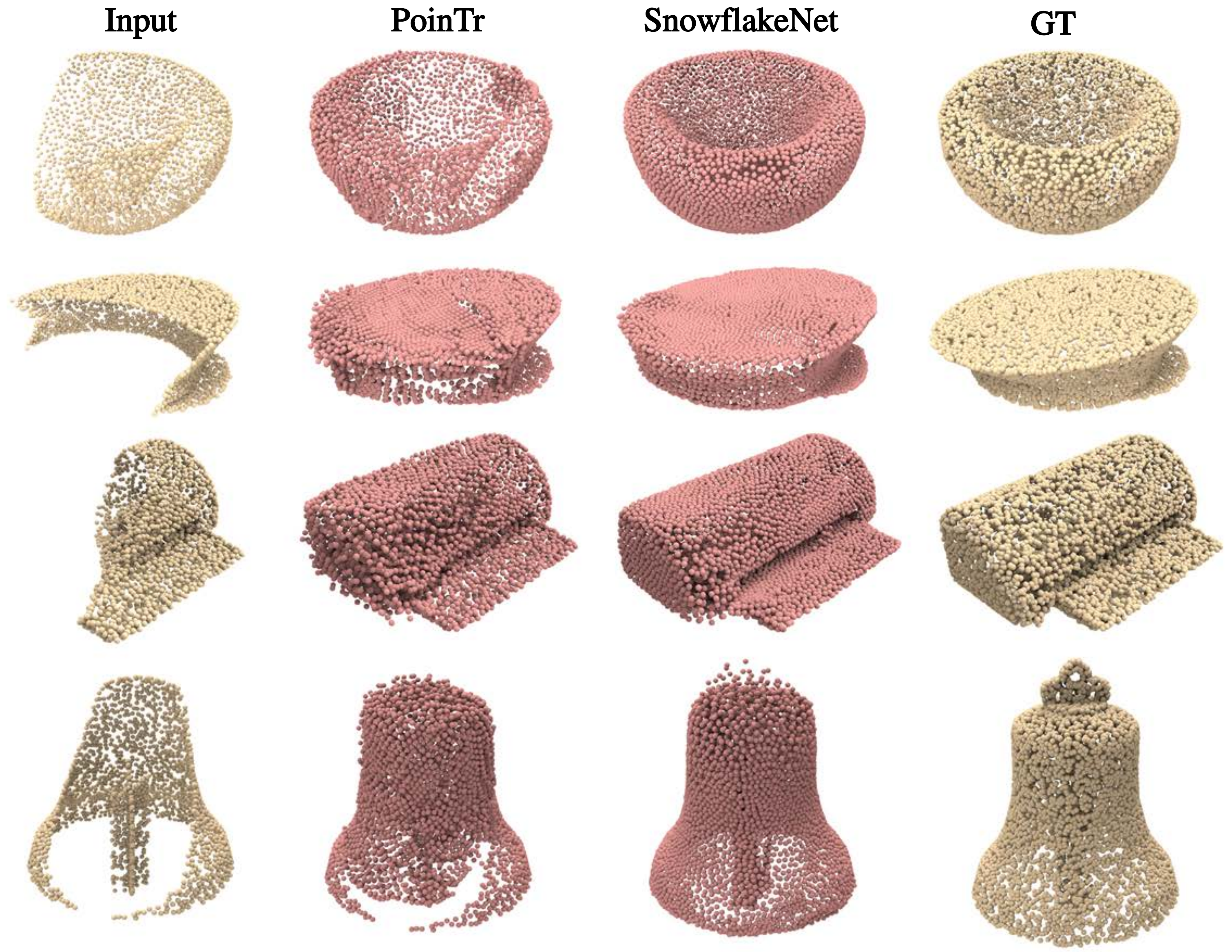}
    \caption{Visual comparison of objects from novel categories in the ShapeNet-34/21 \cite{Yu_2021_ICCV} dataset.}
    \label{fig:shapenet-21}
\end{center}
\vspace{-2em}
\end{figure}

\subsection{Evaluation on the ShapeNet-34/21 Dataset}
\subsubsection{Dataset briefs and evaluation metric}
The ShapeNet-34/21 dataset \cite{Yu_2021_ICCV} is derived from the original ShapeNet \cite{chang2015shapenet} dataset, where 34 observed shape categories are used for training, and the other 21 unseen categories are used for novel shape completion. The 34 categories of observed objects contain 50,165 objects, where those of the train/test split are 46,765 and 3,400 respectively. The unseen 21 categories contain 2,305 objects. To obtain a fair comparison, we follow the same practice of PoinTr \cite{Yu_2021_ICCV} to sample 2,048 points from the object as input and 8,192 points as the ground truth, and the partial point clouds are generated online by removing the furthest $n$ (2,048 to 6,144) points from a randomly selected viewpoint.
To produce complete point clouds with 8,192 points, we set the upsampling factors to $r_1 = 1, r_2 = 4$ and $r_3 = 4$ to keep the output point number the same as other counterparts.

During the evaluation, $L_2$ CD and F-Score \cite{reconstruction_survey, Yu_2021_ICCV} are adopted as the evaluation metrics, where CD can evaluate the similarity between the completion results and the ground truth, and F-Score serves to evaluate the accuracy and completeness of the completed point clouds \cite{zhang2021ganinversion}. According to the removed point number $n$, the completion performance is evaluated under the following different difficulty degrees: \emph{Simple} ($n=2,048$), \emph{Moderate} ($n=4,096$) and \emph{Hard} ($n=6,144$). For each model, 8 fixed viewpoints are set to generate input point clouds. Similar to PoinTr, we present the completion results under different difficulty degrees and the averaged results.

\subsubsection{Quantitative comparison}
Table \ref{table:shapenet34} shows the quantitative comparison between SnowflakeNet and the other counterparts on the ShapeNet-34/21 dataset. The results in Table \ref{table:shapenet34} demonstrate that our method outperforms all the other counterparts in terms of CD on the three difficulty levels, which indicates that SnowflakeNet can handle point clouds of different degrees of incompleteness, including the \emph{Easy, Moderate} and \emph{Hard} levels. Moreover, the results of the unseen categories also proved the generalization ability of SnowflakeNet on novel shape completion. Particularly, compared with PoinTr \cite{Yu_2021_ICCV}, SnowflakeNet reduces the CD-Avg of unseen categories by 5.4, which is 26.3\% lower than that of PoinTr. Meanwhile, SnowflakeNet also achieves competitive results in terms of F1 score, which is very close to that of PoinTr and significantly better than those of the other methods.

\subsubsection{Visual comparison}
In Fig. \ref{fig:shapenet-21}, we show the visual comparison of novel shape completion, where the objects are from unseen categories during training. The completion results in Fig. \ref{fig:shapenet-21} show that SnowflakeNet can reveal the underlying complete shapes while maintaining a visually uniform point distribution, which can be found in the bowl and the copy machine instances.

\subsection{Extension to Real-World Scenarios}
To evaluate the generalization ability of SnowflakeNet on real-world scenarios, we conduct experiments on the KITTI benchmark \cite{geiger2013vision_kitti, yuan2018pcn} and the ScanNet \cite{dai2017scannet} chairs.

\begin{table}[h]\small
\tabcolsep=0.1cm
\centering
\caption{Quantitative comparison on the KITTI benchmark in terms of fidelity and MMD metrics.}
\resizebox{\linewidth}{!}{\begin{tabular}{l|ccccc}
\toprule
   &PFNet \cite{huang2020pf}  & CRN \cite{wang2020cascaded}    & GRNet \cite{xie2020grnet}  & PoinTr \cite{Yu_2021_ICCV}   &Ours    \\ 
\midrule
Fidelity & 1.137 & 1.023 & 0.816 & \textbf{0.000} & 0.034 \\
MMD & 0.792 & 0.872 & 0.568 & 0.526 & \textbf{0.407} \\
\bottomrule
\end{tabular}}
\label{table:kitti_compare}
\end{table}

\begin{figure}[h]
\begin{center}
    \centering
    \includegraphics[width=\linewidth]{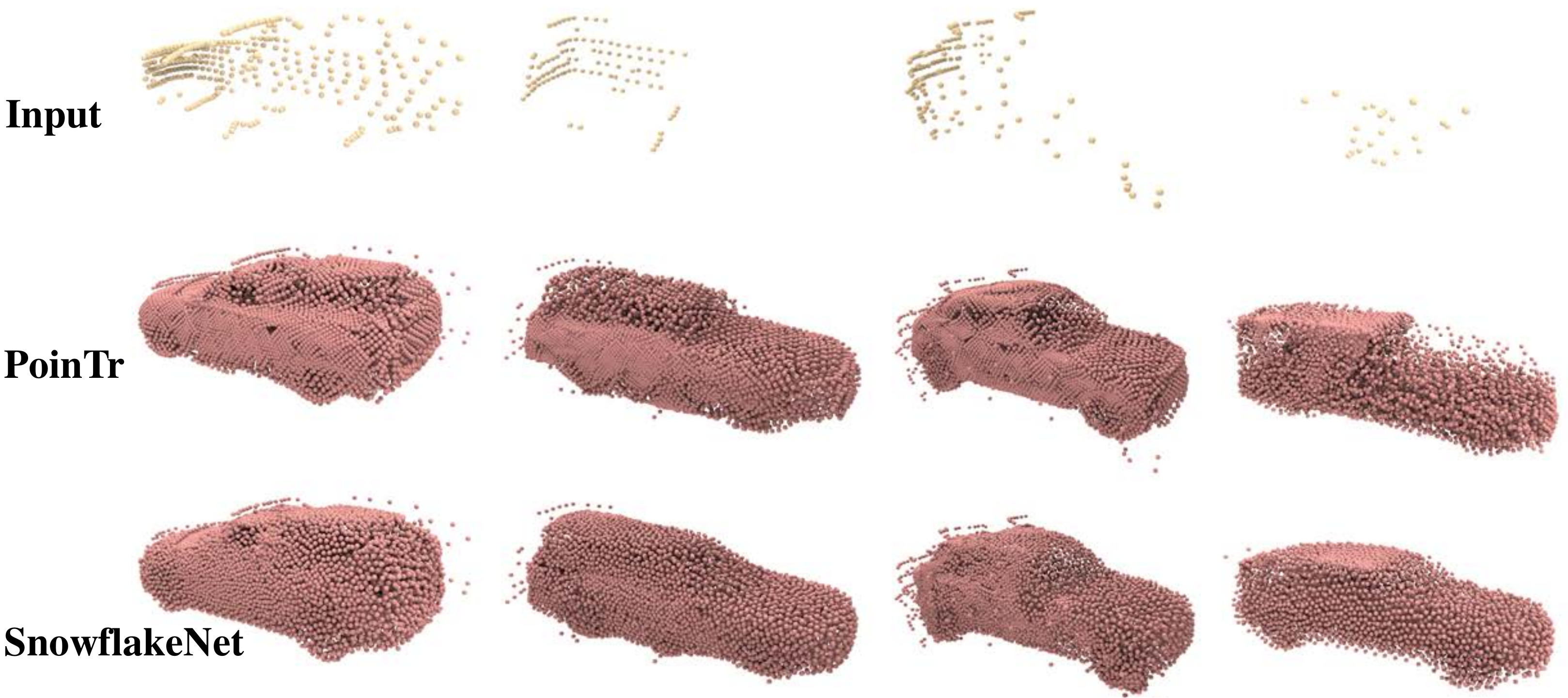}
    \caption{Visual comparison on the KITTI benchmark.}
    \label{fig:kitti}
\end{center}
\vspace{-2em}
\end{figure}

\subsubsection{Extension to the KITTI benchmark}
We test the performance of SnowflakeNet on the KITTI benchmark. Following the same practice of \cite{Yu_2021_ICCV}, we finetune our trained model on ShapeNet cars \cite{chang2015shapenet} and evaluate the performance on the KITTI benchmark. The quantitative comparison with state-of-the-art methods is shown in Table \ref{table:kitti_compare}, where fidelity and minimal matching distance (MMD) \cite{yuan2018pcn} are adopted as the evaluation metrics. The visual comparison is shown in Fig. \ref{fig:kitti}.

\subsubsection{Extension to ScanNet chairs}
\begin{figure}[h]
\begin{center}
    \centering
    \includegraphics[width=\linewidth]{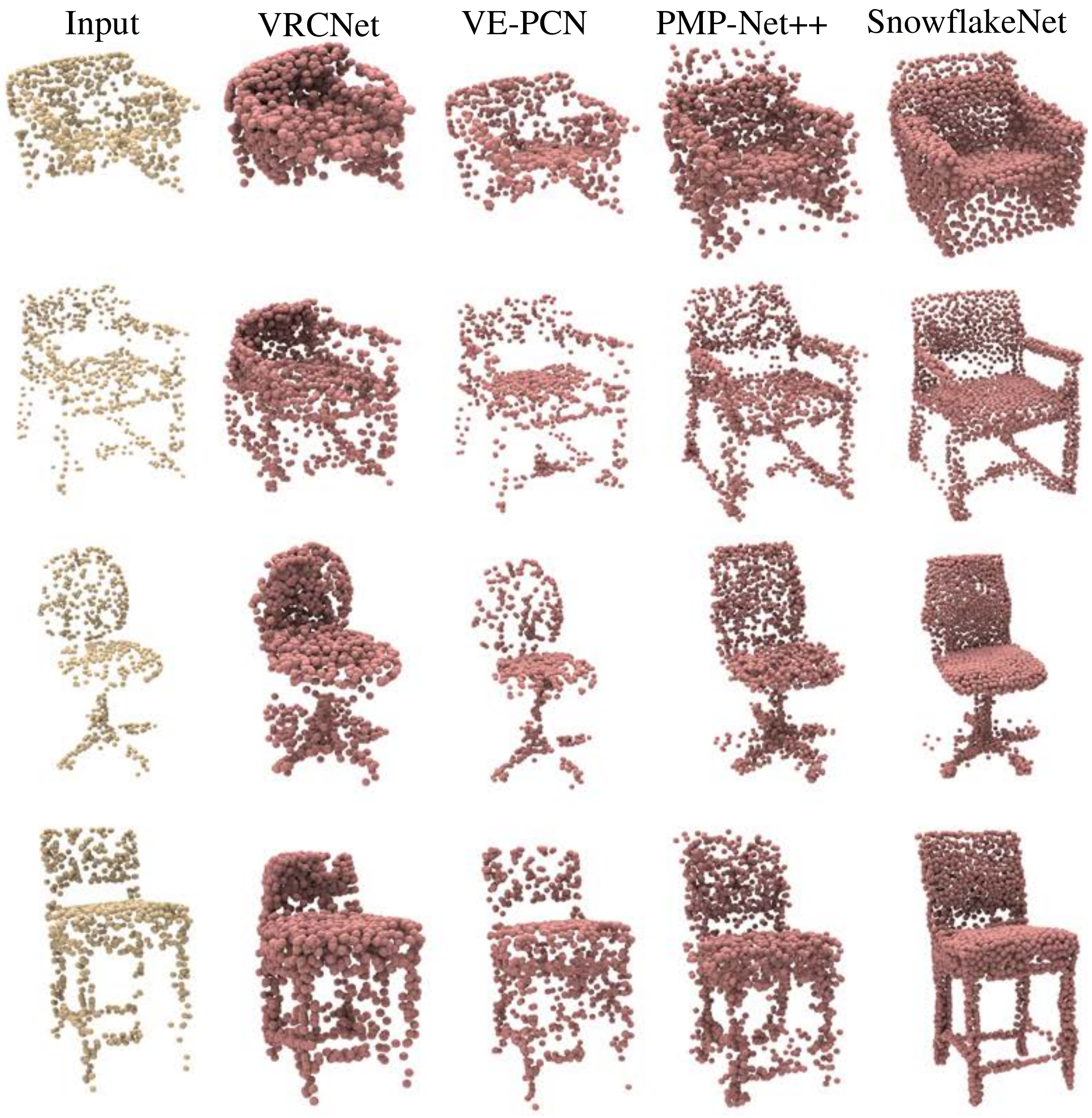}
    \caption{Visual comparison of the chairs in the ScanNet dataset.}
    \label{fig:scannet}
\end{center}
\vspace{-2em}
\end{figure}

%\subsubsection{Extension to ScanNet chairs}
%To evaluate the generalization ability of SnowflakeNet on real world scenario, 
To further evaluate the performance of sparse point cloud completion in the real-world scenario, we use the pre-trained model of SnowflakeNet on the Completion3D dataset and evaluate its performance on the chair instances in the ScanNet dataset without fine-tuning. We compare SnowflakeNet with GRNet\cite{xie2020grnet}, VRCNet \cite{Pan_2021_CVPR_VRCNet}, VE-PCN \cite{Wang_2021_ICCV_VEPCN}, and PMP-Net++\cite{wen2022pmp++} and use their pre-trained models for testing. The quantitative comparison is shown in Table \ref{table:scannet_compare} and the visual comparison is shown in Fig. \ref{fig:scannet}. 
%From Fig. \ref{fig:scannet}, we ca find that even for the shapes that are highly noisy and incomplete, SnowflakeNet completes shapes with less noise than other counterparts, which benefits from the coherent multi-step splittings conducted by consecutive SPDs. 
%And in Fig. \ref{fig:scene}, we provide more visualization results on ScanNet chairs in the same scene.

% \begin{figure*}[t]
% \begin{center}
%     \centering
%     \includegraphics[width=0.95\linewidth]{fig/scene.pdf}
%     \caption{More visualization of completion results on ScanNet chairs.}
%     \label{fig:scene}
% \end{center}
% \end{figure*}

\begin{table}[h]\small
\tabcolsep=0.1cm
\centering
\caption{Quantitative comparison on ScanNet chairs in terms of fidelity and MMD.}
\resizebox{\linewidth}{!}{\begin{tabular}{l|ccccc}
\toprule
   &GRNet \cite{xie2020grnet}  & VRCNet \cite{Pan_2021_CVPR_VRCNet}    & VE-PCN \cite{Wang_2021_ICCV_VEPCN}  & PMP-Net++\cite{wen2022pmp++}   &Ours    \\ 
\midrule
Fidelity & 0.36 & 1.53 & \textbf{0.10} & 0.82 & 0.12 \\
MMD & 6.62 & 10.44 & 11.32 & \textbf{6.22} & 6.29 \\
\bottomrule
\end{tabular}}
\label{table:scannet_compare}
\end{table}

\subsection{Ablation Studies}
We justify the effectiveness of each part of SnowflakeNet. For convenience, we conduct ablation studies on the four categories of the Completion3D dataset. By default, all the experimental settings remain the same as in Section \ref{sec:compleiton3d}.

\begin{table}[h]\small
\centering
\caption{Effect of the skip-transformer.}
\begin{tabular}{l|c|cccc}
\toprule
Methods &avg.  &Couch    &Chair  &Car   &Lamp    \\ 
\midrule
Self-att &8.89	&6.04 &10.9	&9.42	&9.12	\\
No-att 	&9.30	&6.15 &11.2	&10.4	&9.38	\\
No-connect &9.39	&6.17 &11.3	&10.5	&9.51	\\
Full 	&\textbf{8.48} &\textbf{5.89} &\textbf{10.6}	&\textbf{9.32}	&\textbf{8.12}\\
\bottomrule
\end{tabular}
\label{table:st_analysis}
\vspace{-1em}
\end{table}

\subsubsection{Effect of the skip-transformer} To evaluate the effectiveness of skip-transformer used in SnowflakeNet, we develop three network variations as follows. (1) The \emph{Self-att} variation replaces the transformer mechanism in skip-transformer with the self-attention mechanism, where the input is the point features of the current layer. (2) The \emph{No-att} variation removes the transformer mechanism from skip-transformer, where the features from the previous layer of SPD are directly added to the features of the current SPD layer. (3) The \emph{No-connect} variation removes the whole skip-transformer from the SPD layers, and thus, no feature connection is established between the SPD layers. The experimental results are shown in Table \ref{table:st_analysis}. In addition, we denote the original version of SnowflakeNet as \emph{Full} for a clear comparison with the performance of each network variation. From Table \ref{table:st_analysis}, we can find that the Full model achieves the best performance among all compared network variations. The comparison between the No-connect model and the Full model verifies the advantage of the skip-transformer between SPD layers, and the comparison between the No-att model and the Full model further proves the effectiveness of transformer mechanism to learn shape contexts in local regions. Moreover, the comparison between the Self-att model and the No-att model shows that the attention-based mechanism can also contribute to the completion performance.

\subsubsection{Effect of each part in SnowflakeNet} To evaluate the effectiveness of each part in SnowflakeNet, we define four different network variations as follows. (1) The \emph{Folding-expansion} variation replaces the point-wise splitting operation with the folding-based feature expansion method \cite{yang2018foldingnet}, where the features are duplicated several times and concatenated with a 2-dimensional codeword, in order to increase the number of point features. (2) The \emph{$\rm E_{\rm PCN}$+SPD} variation 
employs the PCN encoder and our SnowflakeNet decoder. (3) The \emph{w/o preservation loss} variation removes the partial matching loss. (4) The \emph{PCN-baseline} is the performance of the original PCN method \cite{yuan2018pcn}, which is trained and evaluated under the same settings as our ablation study. In Table \ref{table:pw_expansion_analysis}, we report the results of the four network variations along with the default network denoted as \emph{Full}. By comparing $\rm E_{PCN}$+SPD with PCN-baseline, we can find that our SPD with skip-transformer-based decoder can potentially be applied to other simple encoders, and achieves significant improvement. By comparing the Folding-expansion with the Full model, the better performance of the Full model proves the advantage of point-wise splitting operation over the folding-based feature expansion methods. By comparing the w/o preservation loss model with the Full model, we find that the preservation loss can slightly improve the average performance of SnowflakeNet, but has different effects on different categories.

\begin{table}\small
\centering
\caption{Effect of each part in SnowflakeNet.}
\resizebox{\linewidth}{!}{\begin{tabular}{l|c|cccc}
\toprule 
Methods &avg.  &Couch    &Chair  &Car   &Lamp \\ 
\midrule
Folding-expansion &8.80 &8.40 &10.80 &5.83 &10.10\\
$\rm E_{\rm PCN}$+SPD	&8.93 	&9.06 &11.30  &6.14	&9.23	\\
w/o preservation loss &8.50 &8.72 &\textbf{10.6} &\textbf{5.78} &\textbf{8.9}\\
PCN-baseline &13.30 	&11.50   &17.00 &6.55	&18.20		\\
Full 	&\textbf{8.48} &\textbf{8.12} &10.6	&5.89	&9.32	\\
\bottomrule
\end{tabular}}
\label{table:pw_expansion_analysis}
\vspace{-1em}
\end{table}

\subsubsection{The effect of the splitting strategy}

\begin{table}[h]\small
\centering
\caption{Effect of the splitting strategy.}
\begin{tabular}{l|c|cccc}
\toprule
Splitting strategy &avg.  &Couch    &Chair  &Car   &Lamp    \\ 
\midrule
one-step          &9.53   &6.03	&11.0	&9.94	&9.26 \\
two-steps       &8.64   &6.00	&10.3	&9.62	&8.66\\
baseline  &\textbf{8.48}   &\textbf{5.89}	&10.6	&\textbf{9.32}	&\textbf{8.12}\\
three-steps    &8.66   &6.04	&\textbf{10.0}	&9.53	&9.02\\
\bottomrule
\end{tabular}
\label{table:r_analysis}
\end{table}

SPD is flexible in increasing the point number. When $r_i=1$ ($r_i$ is the upsampling factor of the $i$-th SPD), it serves to move the point from the previous step to a better position; when $r_i > 0$, it splits a point by a factor of $r_i$. In this section, we analyze the influence of different point splitting strategies on the performance of SnowflakeNet, and the other experimental settings are the same as those in the ablation study. As shown in Table \ref{table:r_analysis}, we additionally test three different splitting strategies. The one-step strategy adopts a single SPD (upsampling factor is 4) to split $\mathcal{P}_0$ (512 points) to $\mathcal{P}_1$ (2048 points). The two-step strategy adopts two SPDs (upsampling factors are both equal to 2) and produces three point clouds $\mathcal{P}_0, \mathcal{P}_1$, and $\mathcal{P}_2$ with a size of $512 \times 3, 1024 \times 3$, and $2048 \times 3$, respectively. For the three-step strategy (three SPDs of upsampling factor 2), we particularly set $N_c = N_0 = 256$ (see Section \ref{sec:method:overview}) so that it outputs 4 point clouds of the sizes $256 \times 3, 512 \times 3, 1024 \times 3$, and $2048 \times 3$. Note that the baseline is two-step splitting with an additional SPD (upsampling factor equals 1), which serves to rearrange the initial point positions. By comparing the one-step strategy with the other strategies, we can find that multiple steps of splitting boost the performance of SnowflakeNet significantly (in terms of average CD), this should credit to the collaboration between SPDs. By comparing two-step strategy with three-step strategy, we can find that the two-step splitting suffices for generating a sparse point cloud with 2048 points (not necessarily for a dense one with more points), and extra SPDs may not improve the performance but will increase the computational burden. By comparing the two-step strategy with the baseline, we can find that the additional SPD that adjusts the initial point positions can facilitate the point splitting process.

\subsubsection{Effect of neighbor number $k$}
As mentioned in Section \ref{sec:skip_transformer}, the skip-transformer aggregates local geometric pattern among the $k$ nearest neighbors. Therefore, Table \ref{table:ablation_knn} shows the effect of $k$, which is denoted as \emph{Full}. Additionally, since \emph{Self-att} variation (see Table \ref{table:st_analysis}) also adopts the $k$-NN strategy to conduct self-attention, we also explore its performance. The results in Table \ref{table:ablation_knn} indicate that under different neighbor numbers $k$, the \emph{Full} model outperforms the \emph{Self-att} variation. Moreover, a small $k$ (4, 8, and 16) cannot help to improve the performance of \emph{Self-att}. In contrast, the performance of the \emph{Full} model can steadily improve as the $k$ increases from 4 to 32, which further proves the necessity and effectiveness of skip-transformer. Furthermore, as $k$ becomes larger, the performance of \emph{Full} will not change dramatically, since an oversized neighborhood could introduce noise and the skip-transformer cannot focus on the local context. In our paper, we set $k$ to 16 to balance the trade-off between performance and inference efficiency.

\begin{table}[h]\small
\centering
\caption{Effect of $k$-NN.}
\begin{tabular}{l|ccccc}
\toprule
\multicolumn{1}{c|}{$k$} & 4 & 8 & 16 & 32 & 64 \\
\midrule
Self-att & 8.83 & 8.71 & 8.89 & 8.64 & 8.61 \\
Full & \textbf{8.69} & \textbf{8.51} & \textbf{8.48} & \textbf{8.37} & \textbf{8.38} \\
\bottomrule
\end{tabular}
\vspace{-1em}
\label{table:ablation_knn}
\end{table}

\subsubsection{Robustness to random noise} 
\begin{figure}[h]
\begin{center}
    \centering
    \includegraphics[width=\linewidth]{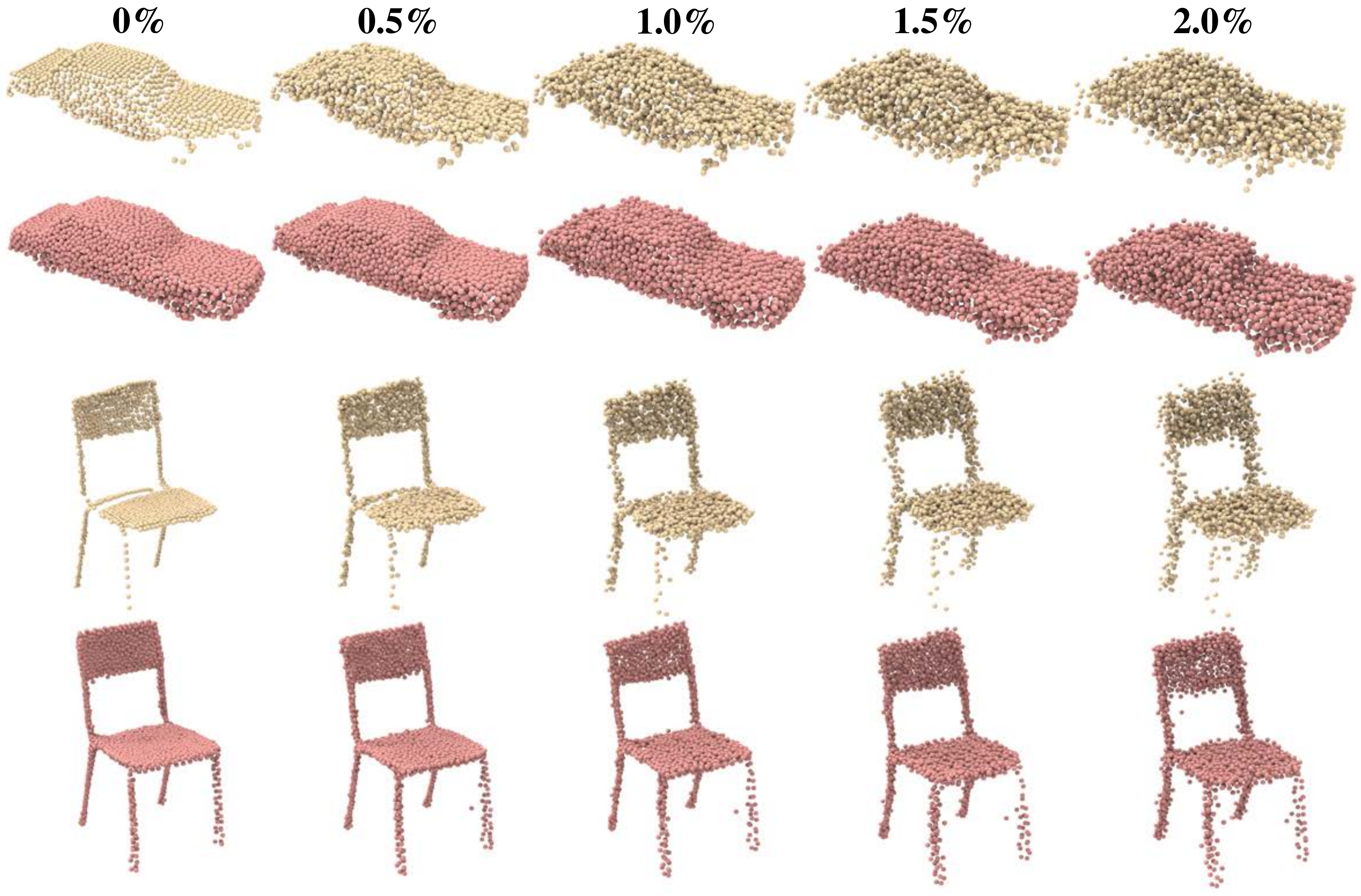}
    \caption{Completion results under different noise levels.}
    \label{fig:noise}
\end{center}
\vspace{-1em}
\end{figure}

We explored the robustness of SnowflakeNet to noise perturbation by adding Gaussian noise to the input point clouds, and the results are shown in Table \ref{table:ablation_noise}. During training, we add zero mean Gaussian noise and change the noise degree by randomly adjusting the standard deviation, which ranges from 0 (0\%) to 0.02 (denoted as 2.0\% in Table \ref{table:ablation_noise}). During the evaluation, we test the performance on five noise levels (0\%, 0.5\%, 1.0\%, 1.5\% and 2.0\%). The quantitative results in Table \ref{table:ablation_noise} show that the random noise during training can hamper the performance on noise-free point clouds (by comparing 0\% with baseline). However, training with noisy point clouds could also significantly improve the robustness to moderate noise degrees, where the network performs stably under noise perturbation levels ranging from 0.5\% to 1.5\%. As the noise level dramatically increases (2.0\%), the quantitative performance will degrade due to the large corruption of partial inputs. We also present visual results under different noise levels in Fig. \ref{fig:noise}, from which we can find that under the 2.0\% noise level, the input point cloud almost lost its underlying structure (see the car instance), but our model can still complete the overall shape and produce plausible point clouds.

\begin{table}[h]\small
\centering
\caption{Robustness to random noise. We add noise of levels ranging from 0\% to 2.0\%, where 2.0\% denotes zero mean Gaussian noise and the standard deviation is set to 0.02.}
\begin{tabular}{l|c|cccc}
\toprule
Noise levels &avg.  &Couch    &Chair  &Car   &Lamp    \\ 
\midrule
baseline  &\textbf{8.48}   &\textbf{5.89}	&\textbf{10.60}	&9.32	&\textbf{8.12}\\
0\% & 8.81 & 6.04 & 11.05 & 9.05 & 9.09 \\
0.5\% & 8.79 & 6.10 & 11.00 & \textbf{8.94} & 9.14 \\
1.0\% & 8.90 & 6.32 & 10.99 & 9.11 & 9.17 \\
1.5\% & 9.37 & 6.65 & 11.59 & 9.48 & 9.78 \\
2.0\% & 10.56 & 7.29 & 13.21 & 10.72 & 11.04 \\

\bottomrule
\end{tabular}
\label{table:ablation_noise}
\vspace{-1em}
\end{table}

\subsubsection{Visualization of point generation process of SPD}
\begin{figure}[t!]
\begin{center}
   \includegraphics[width=\linewidth]{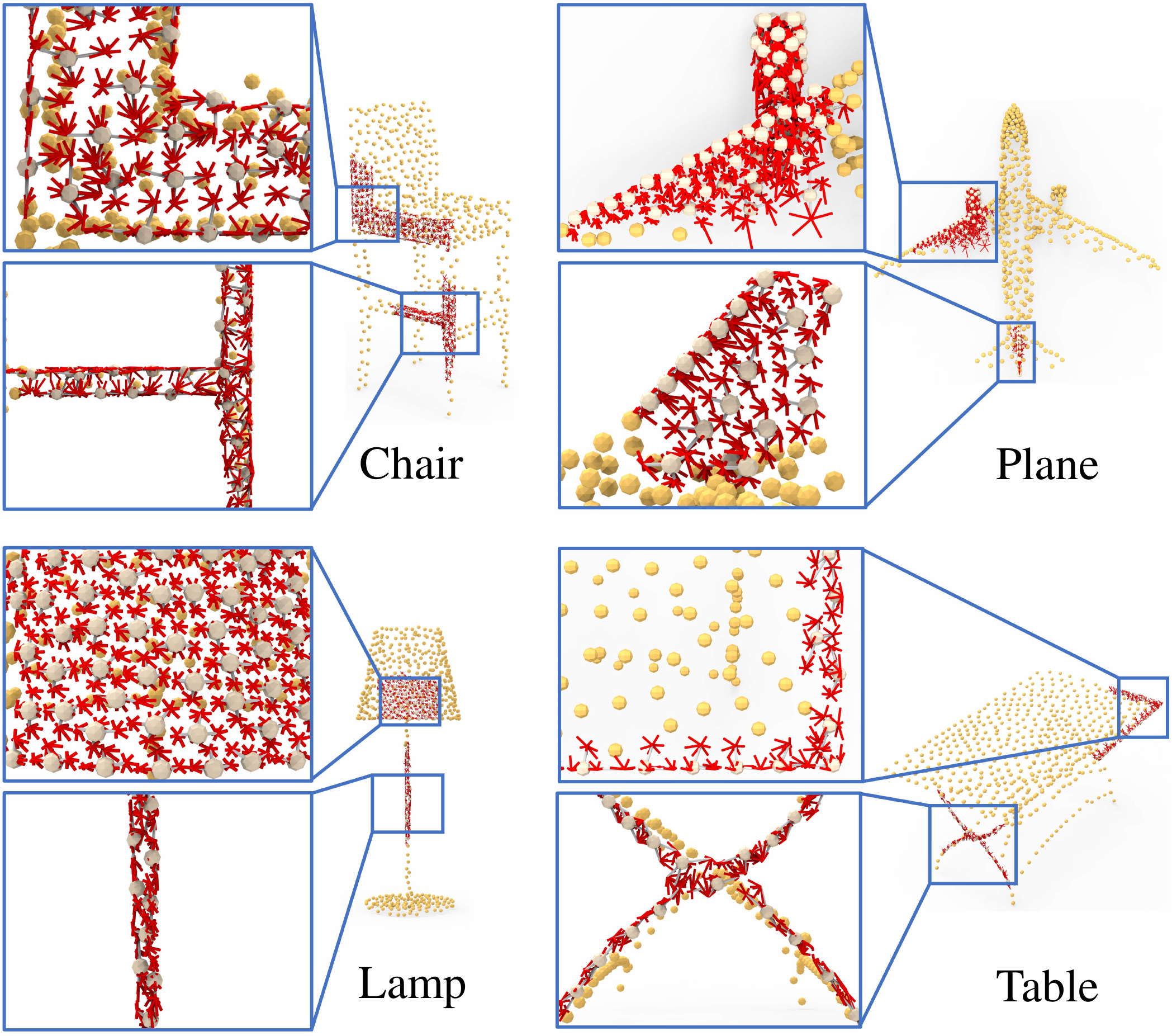}
\end{center}
    \caption{Visualization of snowflake point deconvolution on different objects. For each object, we sample two patches of points and visualize two layers of point splitting together for each sampled point. The gray lines indicate the paths of the point splitting from $\mathcal{P}_1$ to $\mathcal{P}_2$, and the red lines are splitting paths from $\mathcal{P}_2$ to $\mathcal{P}_3$.}
   %\caption{Visualization comparison of point cloud completion on Completion3D dataset. Our SnowflakeNet can produce more smooth surfaces (e.g. car and table) and detailed structures (e.g. boat) compared with the other state-of-the-art point cloud completion methods.}
\label{fig:snowflake}
\vspace{-1em}
\end{figure}

In Fig. \ref{fig:snowflake}, we visualize the point cloud generation process of SPD. We can find that the SPD layers generate points in a snowflake-like pattern. When generating the smooth plane (e.g., chair and lamp in Fig. \ref{fig:snowflake}), we can see the child points are generated around the parent points and are smoothly placed along the plane surface. However, when generating thin tubes and sharp edges, the child points can precisely capture the geometries.

\begin{table*}[h]\small
\centering
\caption{Comparison of point cloud auto-encoding in terms of CD and EMD on ShapeNet. $L_2$ CD and EMD are multiplied by $10^4$ and $10^2$, respectively.}
\begin{tabular}{ll|cccccc}
\toprule
Dataset &Metric &Atlas (S1)\cite{groueix2018atlasnet}  &Altas (P25)\cite{groueix2018atlasnet}  &PointFlow\cite{Yang_2019_ICCV_pointflow} &ShapeGF\cite{ShapeGF} &DPM\cite{luo2021diffusion} &Ours\\ \midrule
\multirow{2}{*}{Airplane}   &CD  &2.00 &1.80 &2.42 &2.10 &2.19 &\textbf{1.51}\\ 
                            &EMD &4.31 &4.37 &3.31 &3.50 &2.90 &\textbf{2.44}\\  \midrule
\multirow{2}{*}{Car}        &CD  &6.91 &6.50 &5.83 &5.47 &5.49 &\textbf{4.68}\\ 
                            &EMD &5.67 &5.41 &4.39 &4.49 &3.94 &\textbf{3.35}\\  \midrule
\multirow{2}{*}{Chair}      &CD  &5.48 &4.98 &6.80 &5.15 &5.68 &\textbf{4.58}\\ 
                            &EMD &5.56 &5.28 &5.01 &4.78 &4.15 &\textbf{3.67}\\  \midrule 
\multirow{2}{*}{ShapeNet}   &CD  &5.87 &5.42 &7.55 &5.73 &5.25 &\textbf{4.20}\\ 
                            &EMD &5.46 &5.60 &5.17 &5.05 &3.78 &\textbf{3.44}\\   

\bottomrule
\end{tabular}
\label{table:generation}
\vspace{-1em}
\end{table*}

\subsection{Point Cloud Auto-Encoding}

The task of point cloud auto-encoding aims to reconstruct a point cloud from its reduced and encoded representation. Since it heavily relies on the generation ability of the decoder, we use the decoding part of SnowflakeNet in point cloud auto-encoding to evaluate the generation ability of snowflake point deconvolution.
\subsubsection{Dataset and evaluation metric}
\textbf{Dataset}. To fairly evaluate the generation ability of SPD, we follow the same experimental settings of DPM \cite{luo2021diffusion} and conduct point cloud auto-encoding on the ShapeNet \cite{chang2015shapenet} dataset. The ShapeNet \cite{chang2015shapenet} dataset contains 51,127 shapes from 55 categories. We use the same training, testing, and evaluation split of DPM \cite{luo2021diffusion}, where the ratios of training, testing and, validation sets are $80\%$, $15\%$, and $5\%$, respectively. To evaluate the generation ability under both simple and difficult settings, we conduct experiments on three categories (i.e., \emph{car}, \emph{airplane} and \emph{chair}) of ShapeNet separately, and also verify the performance of SPD on the entire ShapeNet dataset.
%Four datasets are used in evaluation, which include three categories (i.e. \emph{car}, \emph{airplane} and \emph{chair}) in ShapeNet and the whole ShapeNet dataset. 

\noindent\textbf{Evaluation metric}. The commonly used Chamfer distance (CD) and Earth Mover's distance (EMD) are adopted as the evaluation metrics.

\subsubsection{Network arrangement and training loss}
\label{section:neta_autoencoding}
\textbf{Network arrangement}. The auto-encoding task aims to evaluate the reconstruction ability of our decoder. We follow DPM and replace our feature extractor with a simple PointNet \cite{qi2017pointnet} encoder, so that the encoding settings are the same as those of other methods. In addition, the ground truth shapes used in auto-encoding have 2048 points, so we use our seed generator and two stacked SPDs (each has an upsampling factor of 2) as the decoder. In the seed generator, we set $N_c$ to 512 and take $\mathcal{P}_c$ as the output of the seed generator, where $\mathcal{P}_c$ and $\mathcal{P}_0$ are the same point cloud. This arrangement ensures that the generation of point cloud relies only on the latent code extracted from the encoder.

\noindent\textbf{Training loss}. We use $L_2$ Chamfer distance (CD) in Eq. (\ref{eq:cd}) and Earth Mover's distance (EMD) in Eq. (\ref{eq:emd}) as our reconstruction loss, which is given as follows:
\begin{equation}\small
\mathcal{L}_{recon} = \sum_{i \in \{0, 2\}} \mathcal{L}_{\mathrm{CD}_2} (\mathcal{P}_i, \mathcal{P}_i^{'}) + \mathcal{L}_{\mathrm{EMD}}(\mathcal{P}_i, \mathcal{P}_i^{'}),
\label{eq:loss_recon}
\end{equation}
where $\mathcal{P}_i^{'}$ is the downsampled ground truth shape that has the same point density of $\mathcal{P}_i$.

\begin{figure}[h]
\begin{center}
    \centering
    \includegraphics[width=\linewidth]{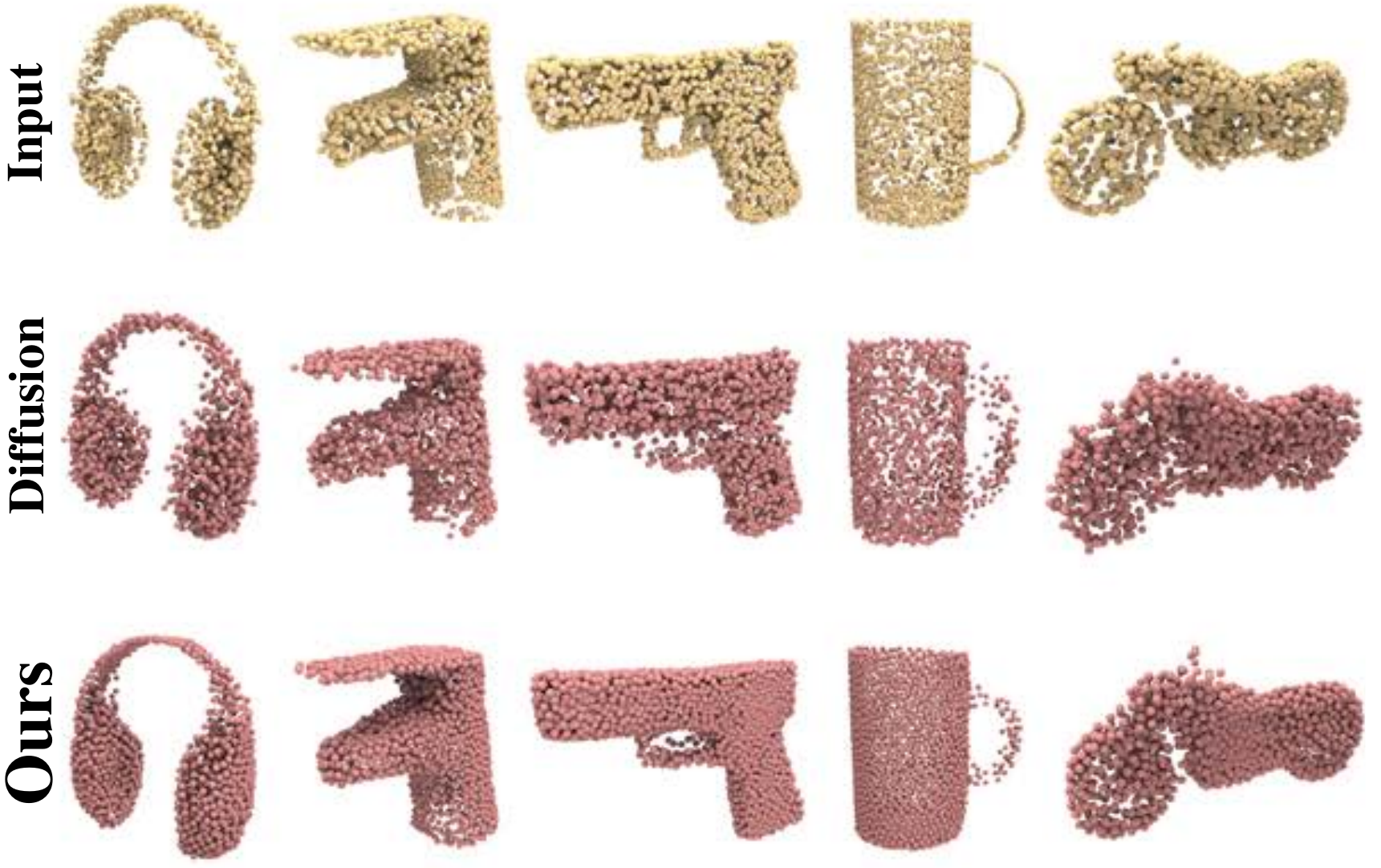}
    \caption{Visual comparison of reconstructed point clouds between different auto-encoders.}
    \label{fig:generation}
\end{center}
\vspace{-1em}
\end{figure}

\begin{figure}[h]
\begin{center}
    \centering
    \includegraphics[width=\linewidth]{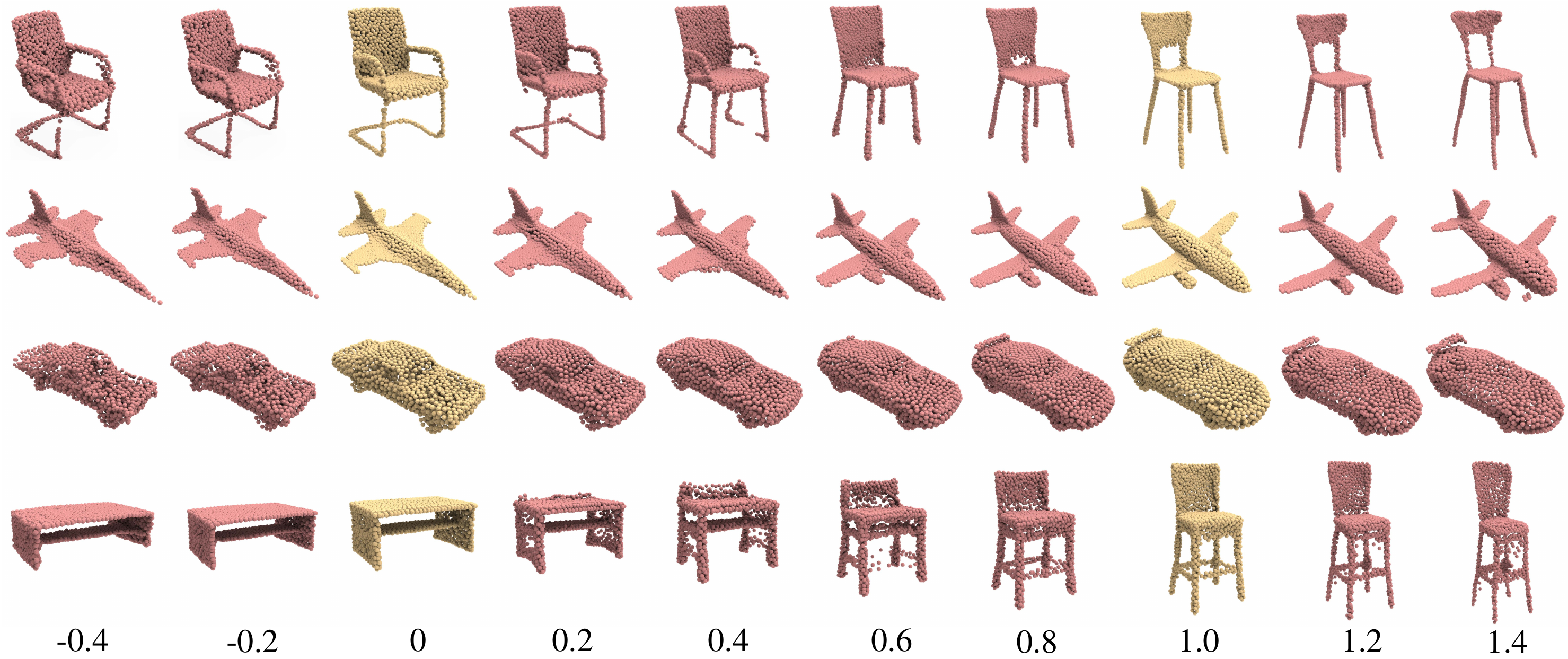}
    \caption{Latent space interpolation and extrapolation. The interpolation step is 0.2, the source shapes are in yellow (denoted as ``0'' and ``1.0''), and the interpolated and extrapolated shapes are in red.}
    \label{fig:interpolation}
\end{center}
\vspace{-1em}
\end{figure}

\subsubsection{Quantitative comparison}
The quantitative comparison is given in Table \ref{table:generation}. Even though only two SPDs are used in the decoder, our method still achieves the best performance among the compared counterparts in terms of both $L_2$ CD and EMD. Especially on the entire ShapeNet dataset, SnowflakeNet reduces the average CD by 1.05, which is 20\% lower than that of DPM\cite{luo2021diffusion} (5.25 in terms of average CD). At the same time, our method also outperforms the other methods across all categories in terms of per-category CD and EMD. Therefore, the better results on point cloud auto-encoding fully demonstrate the generation ability of the decoder of our SnowflakeNet, especially snowflake point deconvolution.

\subsubsection{Visual comparison}
The qualitative comparison is shown in Fig. \ref{fig:generation}. We visually compare SnowflakeNet with the state-of-the-art generation method DPM. Fig. \ref{fig:generation} shows that SnowflakeNet is able to generate point clouds with more detailed geometric structures, such as the handle of the cup and the motor tires. Meanwhile, the points generated by SnowflakeNet are much more uniform and tend to cover the surface of the shape evenly, which should be credited to the excellent generation ability of SPDs. In addition, we visualize the interpolation and extrapolation between latent codes in Fig. \ref{fig:interpolation}, where the source shapes are in yellow, and the interpolated and extrapolated shapes are in red. Fig. \ref{fig:interpolation} shows that the output of SnowflakeNet is able to transit smoothly according to latent interpolation and extrapolation, even though the transition is between different categories (table and chair, the last row). Moreover, the interpolated and extrapolated shapes can still maintain a visually uniform point distribution, which shows the generation stability of our decoder.

\subsection{Novel Shape Generation}
In addition to point cloud auto-encoding, we also apply our network to point cloud generation to demonstrate its novel shape generation ability.

\begin{table*}[!t]\small
\centering
\caption{Comparison of novel point cloud generation performance in terms of $L_2$ CD and EMD. CD is multiplied by $10^4$ and EMD is multiplied by $10^2$.}
\begin{tabular}{ll|cc|cc|cc|c}
\toprule
\multirow{2}{*}{} &\multirow{2}{*}{}  &\multicolumn{2}{c|}{COV($\mathcal{\%}, \uparrow$)} &\multicolumn{2}{c|}{MMD($\downarrow$)}     &\multicolumn{2}{c|}{1-NNA($\mathcal{\%}, \downarrow$)}   &JSD($\downarrow$)\\ \cmidrule{3-9}
Shape                      &Model                       &CD &\multicolumn{1}{c|}{EMD} &CD &\multicolumn{1}{c|}{EMD} &CD &\multicolumn{1}{c|}{EMD}  &-  \\\midrule 
\multirow{2}{*}{} &PC-GAN\cite{achlioptas2018learning_pcgan}   &42.17 &13.84 &3.819 &1.810 & 77.59 &98.52 &6.188\\
                          &GCN-GAN\cite{valsesia2018learning_gcngan}  &39.04 &18.62 &4.713 &1.650 & 89.13 &98.60 &6.669\\
                          &TreeGAN\cite{shu20193dtreegan}  &39.37 &8.40 &4.323 &1.953  & 83.86 &99.67 &15.646\\
Airplane                  &PointFlow\cite{Yang_2019_ICCV_pointflow} &44.98 &44.65 &3.688 &1.090 &66.39 &69.36 &1.536\\
                          &ShapeGF\cite{ShapeGF}  &\textbf{50.41} &\textbf{47.12} &3.306 &1.027 &\textbf{61.94} &70.51 &\textbf{1.059}\\
                          &DPM\cite{luo2021diffusion}      &48.71 &45.47 &3.276 &1.061 &64.83 &\textbf{75.12} &1.067\\ \cmidrule{2-9}
                          &\textbf{Ours}     &46.29 &44.98 &\textbf{3.271} &\textbf{1.002} & 69.35 &73.15 &1.630  \\ \midrule
\multirow{2}{*}{} &PC-GAN\cite{achlioptas2018learning_pcgan}    &46.23 &22.14 &13.436   &3.104 &69.67 &100.00 &6.649 \\
                  &GCN-GAN\cite{valsesia2018learning_gcngan}   &39.84 &35.09 &15.354   &2.213 &77.86 &95.80  &21.708 \\
                  &TreeGAN\cite{shu20193dtreegan}   &38.02 &6.77 &14.936   &3.613  &74.92 &100.00 &13.282 \\
Chair             &PointFlow\cite{Yang_2019_ICCV_pointflow} &41.86 &43.38 &13.631   &1.856 &66.13 &68.40  &12.474\\
                  &ShapeGF\cite{ShapeGF}   &48.53 &46.71 &13.175   &1.785 &\textbf{56.17} &\textbf{62.69} &\textbf{5.996}\\
                  &DPM\cite{luo2021diffusion}       &\textbf{48.94} &\textbf{47.52} &\textbf{12.276}   &\textbf{1.784} &60.11 &69.06  &7.797\\ \cmidrule{2-9}
                  &\textbf{Ours}      &48.83 &36.91 &12.742   &1.830 &58.85 &75.08  &7.579\\
                          
\bottomrule
\end{tabular}
\label{table:vae}
\vspace{-1em}
\end{table*}

\subsubsection{Dataset and evaluation metric}
\textbf{Dataset}. For the task of point cloud generation, we follow DPM\cite{luo2021diffusion} and conduct experiments on the ShapeNet dataset. We quantitatively compare our method with several state-of-the-art generative models on the two categories (i.e., \emph{airplane} and \emph{chair}) in ShapeNet. The generated point clouds and the reference point clouds are normalized into a bounding box of $[-1, 1]$.

\noindent\textbf{Evaluation metric}. To evaluate the generation quality of our method, we follow prior works\cite{Yang_2019_ICCV_pointflow, ShapeGF, luo2021diffusion} and employ the coverage score (COV), the minimum matching distance (MMD), the 1-NN classifier accuracy (1-NNA), and the Jenson-Shannon divergence (JSD). The COV score measures whether the generated samples cover all the modes of the data distribution and the MMD score measures the fidelity of the generated samples. The 1-NNA measures the similarity between the distributions of the generated samples and the reference samples. If the accuracy of the 1-NN classifier is closer to 50\% (random guess), and the generation quality is considered to be better. The JSD score measures the similarity between the marginal point distributions of the generated set and the reference set. Meanwhile, we adopt $L_2$ CD and EMD to evaluate the reconstruction quality.

\begin{figure}[h]
\begin{center}
    \centering
    \includegraphics[width=\linewidth]{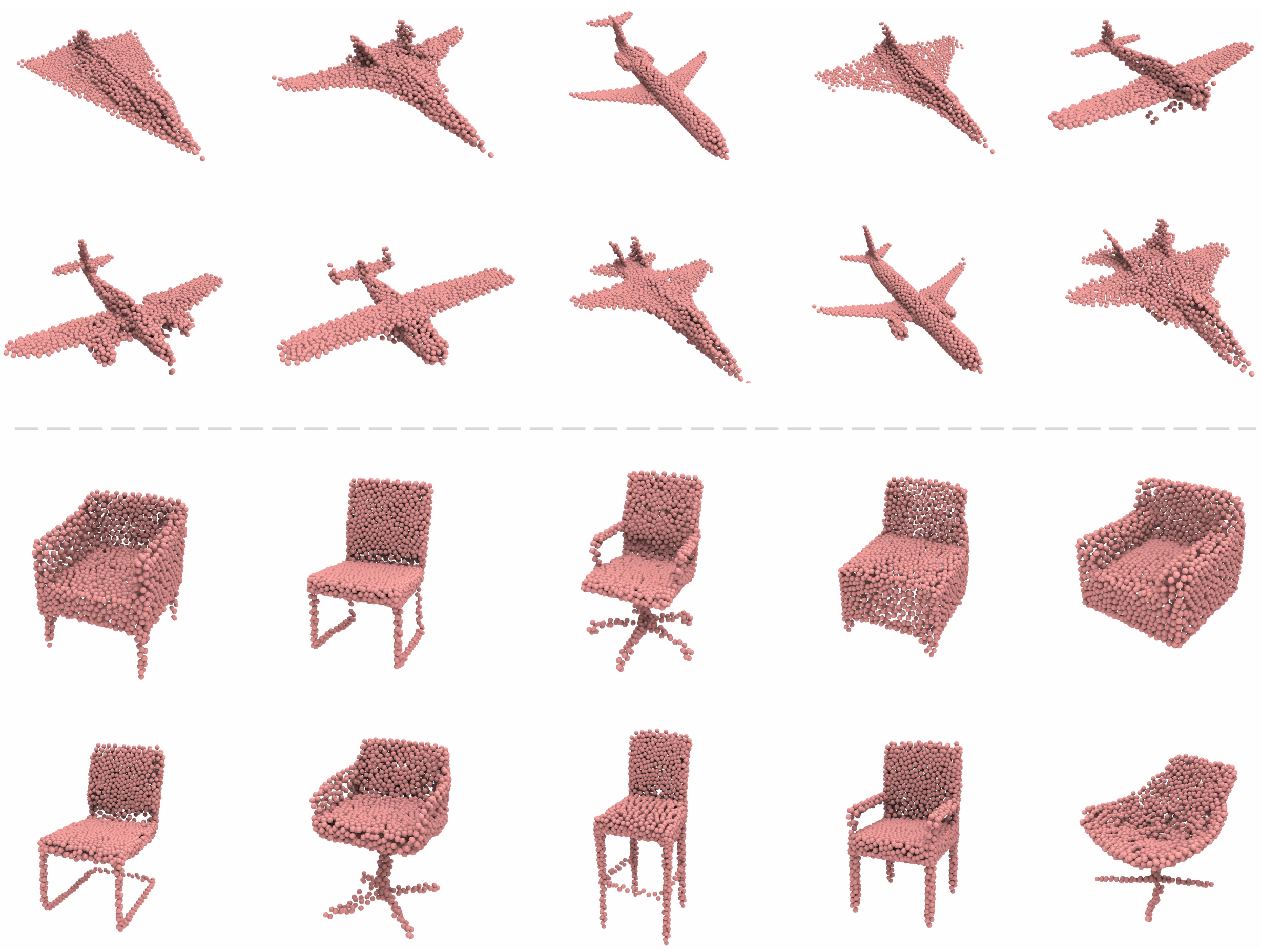}
    \caption{Examples of novel point clouds generated by our model.}
    \label{fig:vae}
\end{center}
\vspace{-7mm}
\end{figure}

\subsubsection{Network arrangement and training loss}
\textbf{Network arrangement}. The network arrangement in this section is the same as point cloud auto-encoding (Section \ref{section:neta_autoencoding}).

\noindent\textbf{Training loss}. To help our decoder to generate novel shapes, we employ normalizing flows\cite{chen2017variational, rezende2015variational, luo2021diffusion} to parameterize the prior distribution $p(\mathbf{f})$, where $\mathbf{f}$ is the latent code extracted by the encoder. The normalizing flow is essentially a stack of affine coupling layers, which provides a trainable bijection $F_\alpha$ that maps an isotropic Gaussian to a complex distribution \cite{luo2021diffusion}. The prior distribution can be computed by: 
\begin{equation}\small
p(\mathbf{f}) = p_{\boldsymbol{w}}(\boldsymbol{w}) \cdot \bigg| \det \frac{\partial F_\alpha}{\partial \boldsymbol{w}} \bigg|^{-1},
\label{eq:prior}
\end{equation}
where $\boldsymbol{w}=F_\alpha^{-1}(\mathbf{f})$, and $p_{\boldsymbol{w}}(\boldsymbol{w})$ is the isotropic Gaussian $\mathcal{N}(\mathbf{0}, I)$. Similar to DPM \cite{luo2021diffusion}, we adopt PointNet\cite{qi2017pointnet} as the architecture for $\mu$ and $\Sigma$ of the encoder $q(\mathbf{f}|\mathcal{P})$. Together with the reconstruction loss in Eq. (\ref{eq:loss_recon}), the training loss for point cloud generation is given as follows:
\begin{equation}
\mathcal{L}_{gen} = D_\mathrm{KL}(q(\mathbf{f}|\mathcal{P}) || p_{\boldsymbol{w}}(\boldsymbol{w}) \cdot \bigg| \det \frac{\partial F_\alpha}{\partial \boldsymbol{w}} \bigg|^{-1}) + \mathcal{L}_{recon},
\label{eq:loss_gen}
\end{equation}
where $D_\mathrm{KL}$ is the KL divergence. To generate a point cloud, we first obtain the latent code $\mathbf{f}$ by drawing $\boldsymbol{w} \sim \mathcal{N}(\mathbf{0}, I)$ and pass $\boldsymbol{w}$ through $F_\alpha$ and then pass the latent $\mathbf{f}$ to our decoder.

\subsubsection{Quantitative and qualitative results}
In Table \ref{table:vae}, we quantitatively compare our methods with the following state-of-the-art generative models: DPM\cite{luo2021diffusion}, ShapeGF\cite{ShapeGF}, PointFlow\cite{Yang_2019_ICCV_pointflow}, TreeGAN\cite{shu20193dtreegan}, GCN-GAN\cite{valsesia2018learning_gcngan}, and PC-GAN\cite{achlioptas2018learning_pcgan}. We follow the same experimental settings as DPM\cite{luo2021diffusion} and all the other results are cited from DPM\cite{luo2021diffusion}. Table \ref{table:vae} shows that our method is highly competitive compared to the latest generative models, such as ShapeGF\cite{ShapeGF} and DPM\cite{luo2021diffusion}, and is significantly better than the other methods. Moreover, we also achieve the best results on the MMD score of the airplane category, which proves the generation ability of our method. In Fig. \ref{fig:vae}, we visually present some point cloud samples generated by our method, from which we find that our method is able to generate novel shapes with good uniformity. Although the latent code is randomly sampled from Gaussian distribution, which leads uncertainty to the decoder, our method can still generate diverse novel shapes while maintaining enriched local geometric structures. Therefore, the visual generation results further demonstrate the excellent generation quality of our snowflake point deconvolution.

\begin{table*}[!t]\small
\centering

\caption{Single image reconstruction on the ShapeNet dataset in terms of per-point $L_1$ Chamfer distance $\times 10^2$ (lower is better).}
\resizebox{18cm}{!}{
\begin{tabular}{l|c|ccccccccccccc}
\toprule
Methods &Average &Plane &Bench &Cabinet &Car &Chair &Display &Lamp &Loud. &Rifle &Sofa &Table &Tele. &Vessel\\ \midrule
3DR2N2\cite{choy20163dR2N2}     &5.41 &4.94 &4.80 &4.25 &4.73 &5.75 &5.85 &10.64 &5.96 &4.02 &4.72 &5.29 &4.37 &5.07\\
PSGN\cite{fan2017psgn}       &4.07 &2.78 &3.73 &4.12 &3.27 &4.68 &4.74 &5.60  &5.62 &2.53 &4.44 &3.81 &3.81 &3.84\\
Pixel2mesh\cite{wang2018pixel2mesh} &5.27 &5.36 &5.14 &4.85 &4.69 &5.77 &5.28 &6.87  &6.17 &4.21 &5.34 &5.13 &4.22 &5.48\\
AtlasNet\cite{groueix2018atlasnet}   &3.59 &2.60 &3.20 &3.66 &3.07 &4.09 &4.16 &4.98  &4.91 &2.20 &3.80 &3.36 &3.20 &3.40\\
OccNet\cite{mescheder2019occupancy}     &4.15 &3.19 &3.31 &3.54 &3.69 &4.08 &4.84 &7.55  &5.47 &2.97 &3.97 &3.74 &3.16 &4.43\\ 
Pix2Vox \cite{xie2019pix2vox} & 4.28 & 3.48 & 4.47 & 4.39 & 3.56 & 4.04 & 4.47 & 5.66 & 5.10 & 3.80 & 4.37 & 4.29 & 3.84 & 4.14 \\
Pix2Vox++ \cite{xie2020pix2vox++} & 4.17 & 3.65 & 4.40 & 3.99 & 3.48 & 3.97 & 4.40 & 5.63 & 4.84 & 3.78 & 4.12 & 4.01 & 3.68 & 4.28 \\
DPM \cite{luo2021diffusion} & 3.76 & 2.64 & 3.56 & 3.46 & 3.23 & 4.15 & 4.35 & 5.18 & 5.14 & 2.41 & 4.15 & 3.71 & 3.42 & 3.52 \\
3DAttriFlow \cite{3DAttriFlow} & 3.02 & 2.11 & 2.71 & 2.66 & 2.50 & 3.33 & 3.60 & 4.55 & 4.16 & 1.94 & 3.24 & 2.85 & 2.66 &  2.96 \\
\midrule
Ours       &\textbf{2.86} &\textbf{1.99} &\textbf{2.54} &\textbf{2.52} &\textbf{2.44} &\textbf{3.13} &\textbf{3.37} &\textbf{4.34}  &\textbf{3.98} &\textbf{1.84} &\textbf{3.09} &\textbf{2.71} &\textbf{2.45} &\textbf{2.80}\\

\bottomrule
\end{tabular}}
\label{table:svr}
\vspace{-1em}
\end{table*}

\subsection{Single Image Reconstruction}
In this section, we extend our network to the task of single image reconstruction (SVR), of which the goal is to reconstruct a point cloud from an image of the underlying object.

\subsubsection{Dataset and evaluation metric}
\label{sec:dset_svr}
\textbf{Dataset}. For the single view reconstruction task, we employ the ShapeNet \cite{chang2015shapenet} dataset containing 43783 shapes from 13 categories. The dataset is split into training, testing, and validation sets by the ratios of 70\%, 20\%, and 10\%, respectively. During training, we track the loss of our method on the validation set to determine when to stop training. The testing set is used for quantitative and qualitative comparison. The baseline methods include 3D-R2N2\cite{choy20163dR2N2}, PSGN\cite{fan2017psgn}, Pixel2Mesh\cite{wang2018pixel2mesh}, AtlasNet\cite{groueix2018atlasnet}, OccNet\cite{mescheder2019occupancy}, and the state-of-the-art methods like Pix2Vox \cite{xie2019pix2vox}, Pix2Vox++ \cite{xie2020pix2vox++}, DPM \cite{luo2021diffusion}, and 3DAttriFlow \cite{3DAttriFlow}. For a fair comparison, we follow \cite{mescheder2019occupancy, 3DAttriFlow} and sample 30k points from the watertight mesh as the ground truth.

\begin{figure}[t]
\begin{center}
    \centering
    \includegraphics[width=\linewidth]{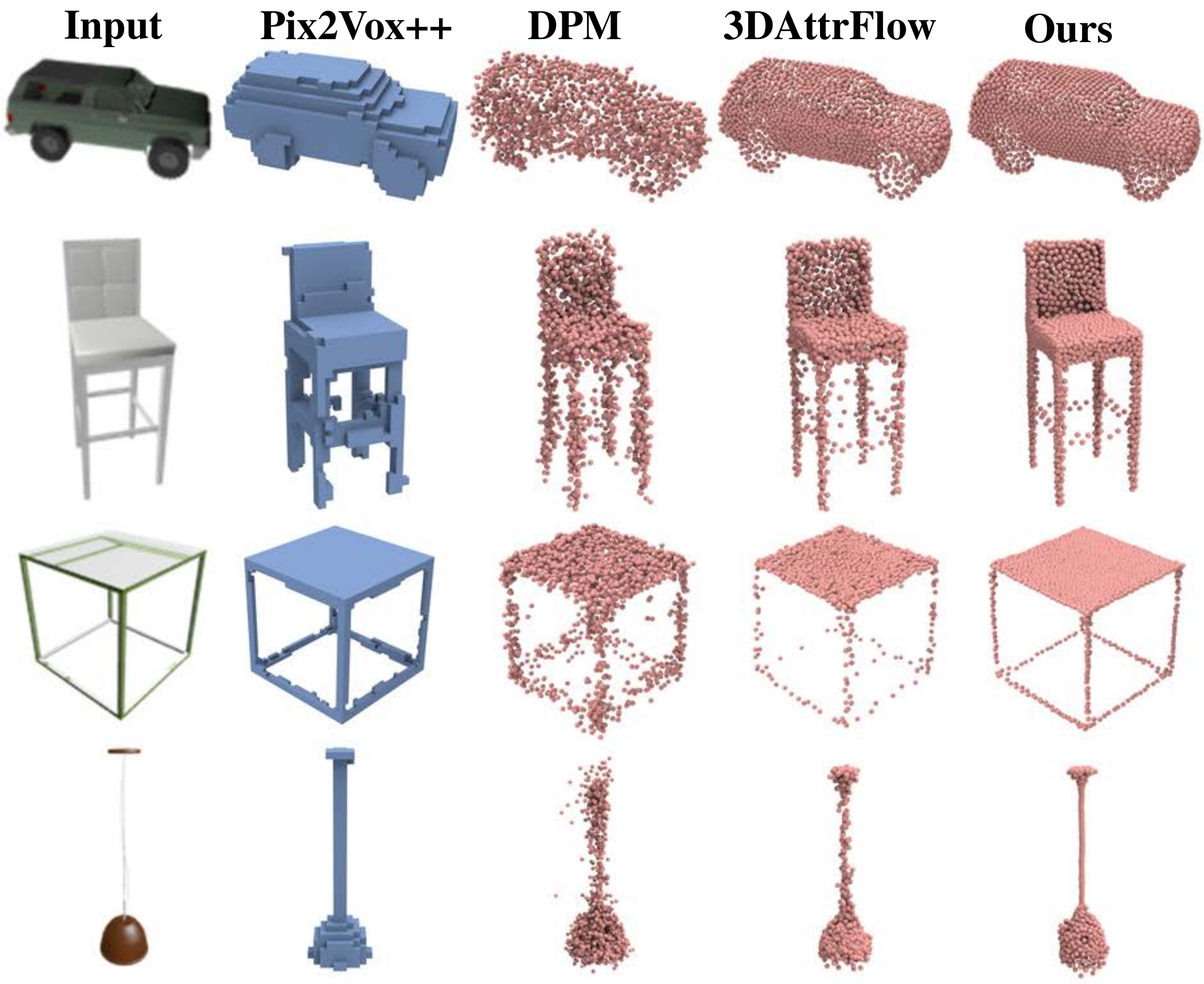}
    \caption{Visual comparison of single image reconstruction. Voxels are in blue, and point clouds are in red.}
    \label{fig:svr}
\end{center}
\vspace{-2em}
\end{figure}

\begin{figure*}[t!]
\begin{center}
   \includegraphics[width=\linewidth]{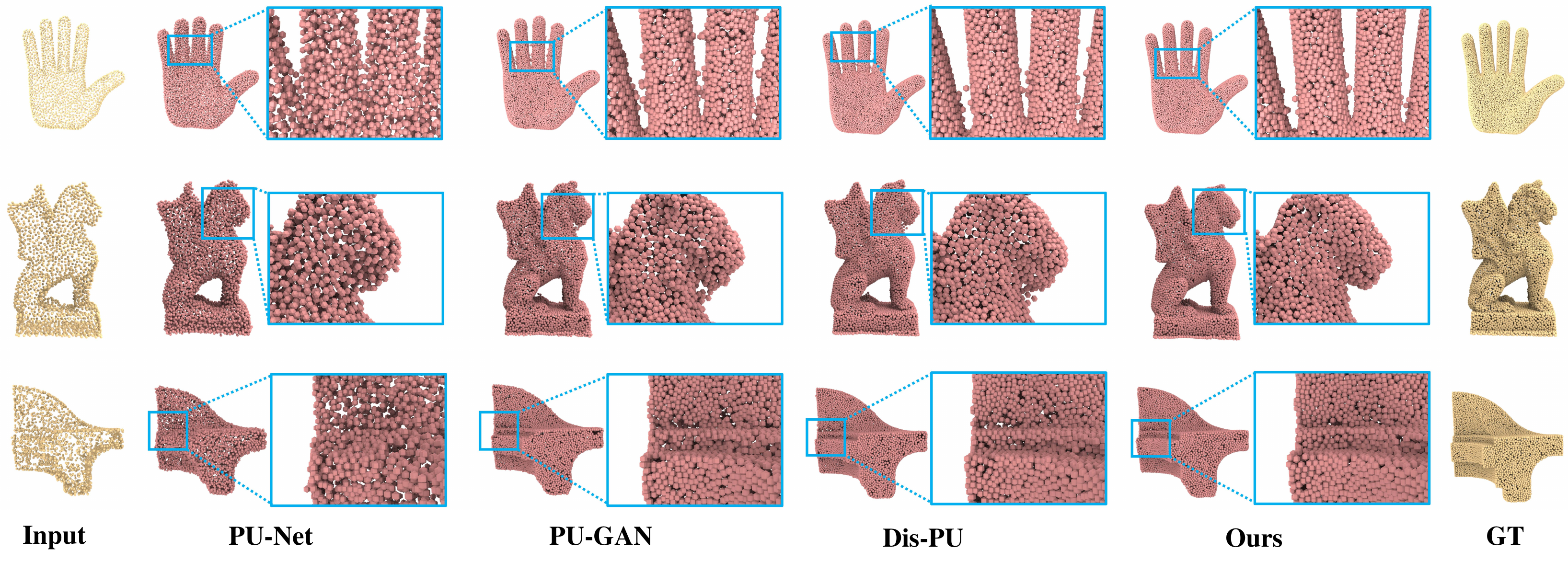}
\end{center}
    \caption{Visual comparison of point cloud upsampling.}
   %\caption{Visualization comparison of point cloud completion on Completion3D dataset. Our SnowflakeNet can produce more smooth surfaces (e.g. car and table) and detailed structures (e.g. boat) compared with the other state-of-the-art point cloud completion methods.}
\label{fig:upsampling}
\vspace{-1em}
\end{figure*}

\noindent
\textbf{Evaluation metric}. We adopt the $L_1$ Chamfer distance (CD) to evaluate the generation quality among all counterparts. In our experiment, we set the output point density of all methods to 2048. The output points of voxel-based (Pix2Vox \cite{xie2020pix2vox++}) and mesh-based (Pixel2Mesh\cite{wang2018pixel2mesh} and OccNet\cite{mescheder2019occupancy}) methods are sampled on their generated mesh, and the sampling strategy is the same as \cite{mescheder2019occupancy}.

\subsubsection{Network arrangement and training loss}
\textbf{Network arrangement}. The network framework of SVR consists of an encoder and a decoder. We use ResNet\cite{he2016deep} (ResNet-18) as the encoder to align with the other baseline methods, and the dimension of the latent code is transformed to 128 by a \emph{linear} layer. For our decoder, the arrangement is the same as auto-encoding (Section \ref{section:neta_autoencoding}).

\noindent\textbf{Training loss}. We use the Chamfer-$L_1$ distance (CD) as the training loss for single image reconstruction, which is given as
\begin{equation}
\mathcal{L}_{\mathrm{SVR}} = \sum_{i \in \{0, 2\}} \mathcal{L}_{\mathrm{CD}_1}(\mathcal{P}_i, \mathcal{P}_i^{i}).
\label{eq:loss_svr}
\end{equation}

\subsubsection{Quantitative comparison}
In Table \ref{table:svr}, we quantitatively compare our method with the baseline methods listed in Section \ref{sec:dset_svr}. The quantitative results in Table \ref{table:svr} demonstrate that our method has the best reconstruction performance among all compared methods. Moreover, our method achieves the best per-category CD across all shape categories, which fully demonstrates the generalization ability of our method.

\subsubsection{Visual comparison}
We visually compare the reconstruction results in Fig. \ref{fig:svr}, which illustrates the best reconstruction quality of our method among all baseline methods. Limited to voxel resolution, the reconstruction results of Pix2Vol++\cite{xie2020pix2vox++} can accurately reconstruct the overall shape but fail to reveal detailed geometries. DPM\cite{luo2021diffusion} produces too much noise and requires additional denoising in post-processing steps. Compared with 3DAttriFlow \cite{3DAttriFlow}, our method is able to reconstruct the underlying object while recovering smooth surfaces, and the SPD enables our decoder to produce shapes with less noise and smoother surfaces.

\subsection{Point Cloud Upsampling}
Given a sparse, noisy, and non-uniform point cloud, the task of point cloud upsampling requires upsampling and generating a dense and uniform point cloud. Because it is also a generation problem, we extend snowflake point deconvolution (SPD) to point cloud upsampling in this section.

\begin{table}[h]\small
\centering
\caption{Quantitative comparison on the point cloud upsampling task in terms of CD $\times 10^3$ and HD $\times 10^3$. FPS stands for farthest point sampling, and RS denotes random sampling.}
\resizebox{\linewidth}{!}{\begin{tabular}{lc|cc|cc}
\toprule
\multirow{2}{*}{Methods} & \multirow{2}{*}{Size} & \multicolumn{2}{c|}{FPS} & \multicolumn{2}{c}{RS} \\
\cmidrule{3-6}

 &  & CD  &HD  & CD & HD\\ \midrule
\multicolumn{1}{l|}{PU-Net\cite{li2018pu}}  &10.1M    &0.473  &4.680 & 0.532 & 5.441\\
\multicolumn{1}{l|}{PU-GAN\cite{li2019pu}} &9.57M   &0.246  &3.011 & \textbf{0.269} & 4.687\\
\multicolumn{1}{l|}{Dis-PU\cite{Li_2021_CVPR}}  &13.2M &0.210  &2.771 & 0.278 & 4.289\\ \midrule
\multicolumn{1}{c|}{\textbf{Ours}}  &\textbf{1.37M}    &\textbf{0.199}  &\textbf{2.701}  &0.270  & \textbf{3.787}\\
\bottomrule
\end{tabular}}
\label{table:upsampling}
\vspace{-1em}
\end{table}

\subsubsection{Dataset and evaluation metric}
\textbf{Dataset}. To compare the upsampling quality of SPD with state-of-the-art methods, we use the benchmark dataset provided by PU-GAN\cite{li2019pu} with 120 training and 27 testing objects. Each training object is cropped into 200 overlapped patches, and there are a total of 24,000 training surface patches. Each surface patch has 1024 points. During training, we randomly sample 256 points as input. For testing objects, we follow \cite{li2019pu} and use Poisson disk sampling to sample 8192 points as the ground truth shape, and 2048 points are further sampled as input. Since the performance of point cloud upsampling is highly sensitive to the quality of input point clouds, we get the inputs under two settings: farthest point sampling (FPS) \cite{qi2017pointnet++} and random sampling. To mitigate the instability brought by random sampling, we repeat the test under random sampling five times and take the average scores as the final results.
%We test the models under two settings: (1) 
%Here the input shapes are sampled from the ground truth point cloud by farthest point sampling (FPS) \cite{qi2017pointnet++} (instead of random sampling), because the size of the testing dataset is relatively small and the quantitative results are highly sensitive to the uniformity of input shapes.

\noindent\textbf{Evaluation metric}. We use the $L_2$ Chamfer distance (CD) and the Hausdorff distance (HD) \cite{berger2013benchmark} as the evaluation metric for a fair comparison with other methods.

\subsubsection{Network arrangement and training loss}
\textbf{Network arrangement}. For the task of point cloud upsampling, four stacked SPDs are used to upsample the sparse point cloud into a dense one. The upsampling factors $r_1, r_2, r_3$, and $r_4$ of the four SPDs are 1, 2, 2, 1, respectively. Because point cloud upsampling requires the output point cloud to be uniform, we add another SPD (of which $r_4$ = 1) to refine the upsampled point cloud.

\noindent\textbf{Training loss}. We use only the $L_2$ Chamfer distance (CD) to train our network, and the training loss is given as follows:
\begin{equation}
\mathcal{L}_{upsampling} = \sum_{i \in \{1, 3, 4\}} \mathcal{L}_{\mathrm{CD}_2}
(\mathcal{P}_i, \mathcal{P}_i^{'} ),
\label{eq:loss_upsampling}
\end{equation}
where $\mathcal{P}_i^{'}$ is the downsampled ground truth shape that has the same point density as $\mathcal{P}_i$.

\subsubsection{Quantitative comparison}
In Table \ref{table:upsampling}, we quantitatively compare our method with the following state-of-the-art point cloud upsampling methods: PU-Net\cite{li2018pu}, PU-GAN\cite{li2019pu}, and Dis-PU\cite{Li_2021_CVPR}. 
FPS stands for farthest point sampling, and RS denotes random sampling. We use the pre-trained models provided by these methods for testing on the same test set. From Table \ref{table:upsampling}, we can find that our method achieves the best performance in terms of both $L_2$ CD and HD under the FPS setting. Even though our method is not trained with repulsion loss\cite{li2018pu} or uniform loss\cite{li2019pu}, our method can recover the underlying surface while maintaining good point uniformity. Meanwhile, under the RS setting, the CD scores of PU-GAN, Dis-PU, and SPD (Ours) are highly close to each other, and we achieve a much better result in terms of HD. Moreover, by comparing the model size with the other methods in terms of parameter number, we find that the model size of our method is significantly smaller than those of the others, which further proves the efficiency of snowflake point deconvolution.

\subsubsection{Visual comparison}
In Fig. \ref{fig:upsampling}, we visually compare our method with the state-of-the-art methods listed in Table \ref{table:upsampling}. From Fig. \ref{fig:upsampling}, we can find that our method can produce the most similar visual results to the ground truth shape. Whether the ground truth shape has complex tiny local structures (the second shape) or simple flat regions (the third shape), our method can preserve a uniform point distribution.

\section{Conclusion, Limitation and Future Work}
In this paper, we propose a novel neural network for point cloud completion, named SnowflakeNet. SnowflakeNet models the generation of complete point clouds as the snowflake-like growth of points in 3D space using multiple layers of snowflake point deconvolution (SPD). By further introducing a skip-transformer in SPD, SnowflakeNet learns to generate locally compact and structured point clouds with highly detailed geometries. We conduct comprehensive experiments on sparse (Completion3D) and dense (PCN) point cloud completion datasets, which shows the superior performance of our method over the current SOTA point cloud completion methods. We further extend SPD to more point cloud generation tasks, such as point cloud auto-encoding, novel point cloud generation, single image reconstruction, and upsampling. The generation ability of Snowflake Point Deconvolution is fully demonstrated by both quantitative and qualitative experimental results.

The main limitation of SnowflakeNet is that the skip-transformer in SPD merely aggregates information from the $k$-nearest neighbors ($k$-NN) of each point. While the $k$-NN strategy enables the SPD to focus on local structure, it also disenables the SPD to capture the long-range shape context. Therefore, when the input point cloud contains too much noise, SPD is unable to arrange point splitting properly. In our opinion, a potential and promising future work to address this problem is to further explore the local and global feature fusion in SPD, so that SPD can adapt well to more complex shape contexts.

Another limitation is that SPD generates the same number of child points for all seed points, despite different numbers of points being needed to represent different parts of the shape. For example, it requires more points to represent the lampshade than the lamppost. It would be ideal if SPD can adaptively split seed points based on the complexity of local shape contexts; this may require significant changes to the network.

\vspace{-1em}

\bibliographystyle{IEEEtran}
\bibliography{SPD}

\begin{IEEEbiography}[{\includegraphics[width=1in,height=1.25in,clip,keepaspectratio]{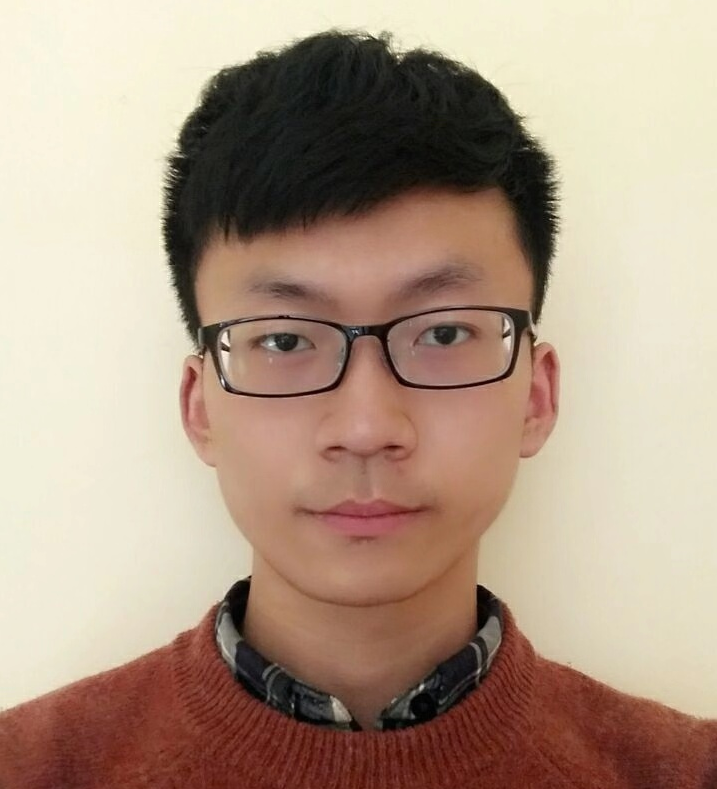}}]{Peng Xiang}
received his B.S. degree in software engineering from Chongqing University, China, in 2019. He is currently a graduate student with the School of Software, Tsinghua University. His research interests include deep learning, 3D shape analysis and pattern recognition.
\end{IEEEbiography}
\vspace{-1em}
\begin{IEEEbiography}[{\includegraphics[width=1in,height=1.25in,clip,keepaspectratio]{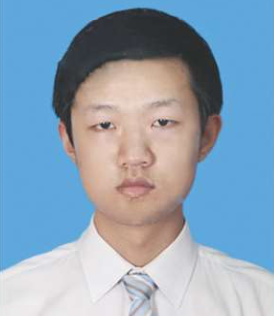}}]{Xin Wen}
received his B.S. degree in engineering management from the Tsinghua University, China, in 2012. He is currently a Ph.D. student with the School of Software, Tsinghua University. His research interests include deep learning, shape analysis and pattern recognition, and NLP.
\end{IEEEbiography}
\begin{IEEEbiography}[{\includegraphics[width=1in,height=1.25in,clip,keepaspectratio]{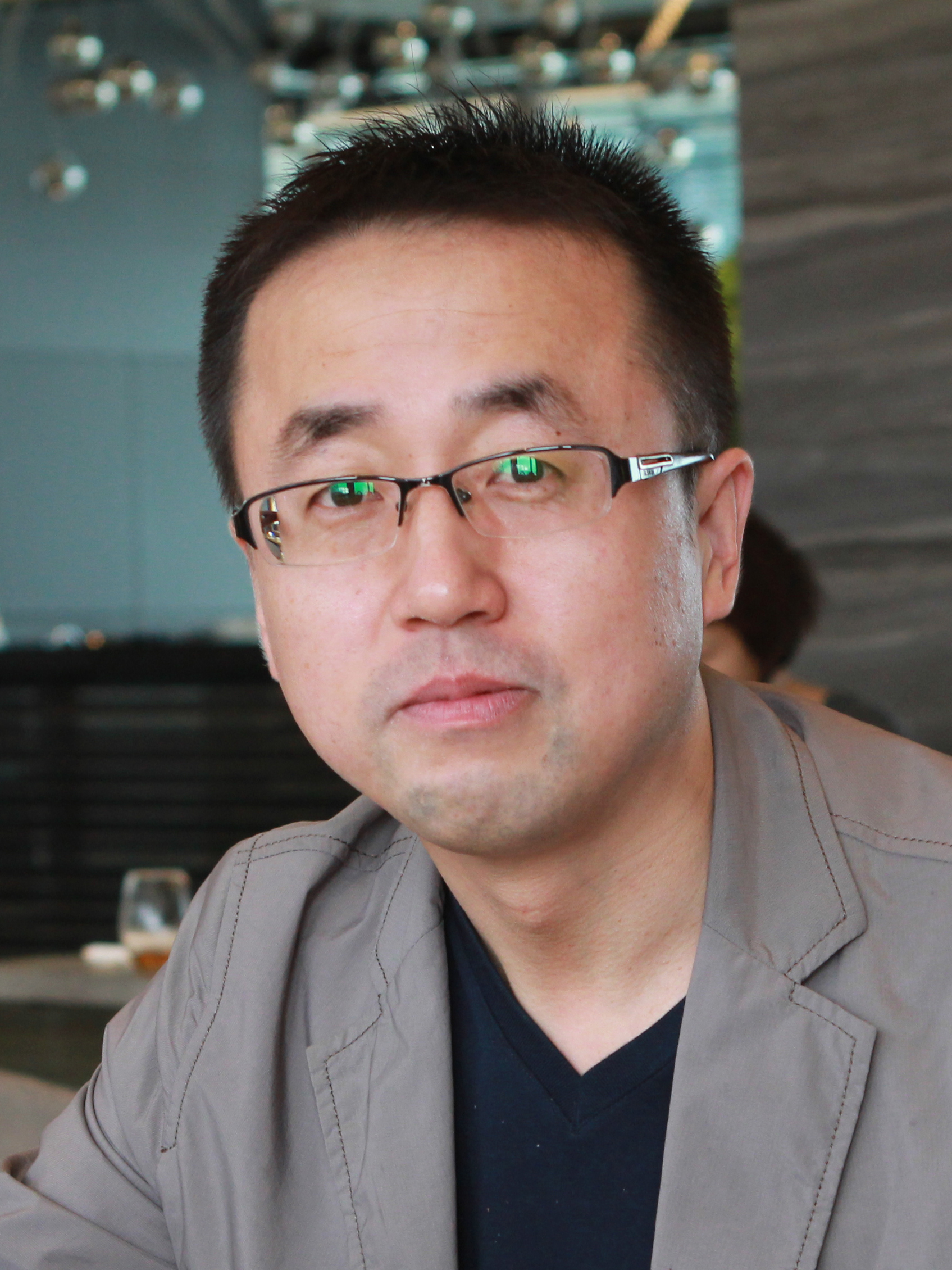}}]{Yu-Shen Liu}
(M'18) received the B.S. degree in mathematics from Jilin University, China, in 2000, and the Ph.D. degree from the Department of Computer Science and Technology, Tsinghua University, Beijing, China, in 2006. From 2006 to 2009, he was a Post-Doctoral Researcher with Purdue University. He is currently an Associate Professor with the School of Software, Tsinghua University. His research interests include shape analysis, pattern recognition, machine learning, and semantic search.
\end{IEEEbiography}
\vspace{-1em}
\begin{IEEEbiography}[{\includegraphics[width=1in,height=1.25in,clip,keepaspectratio]{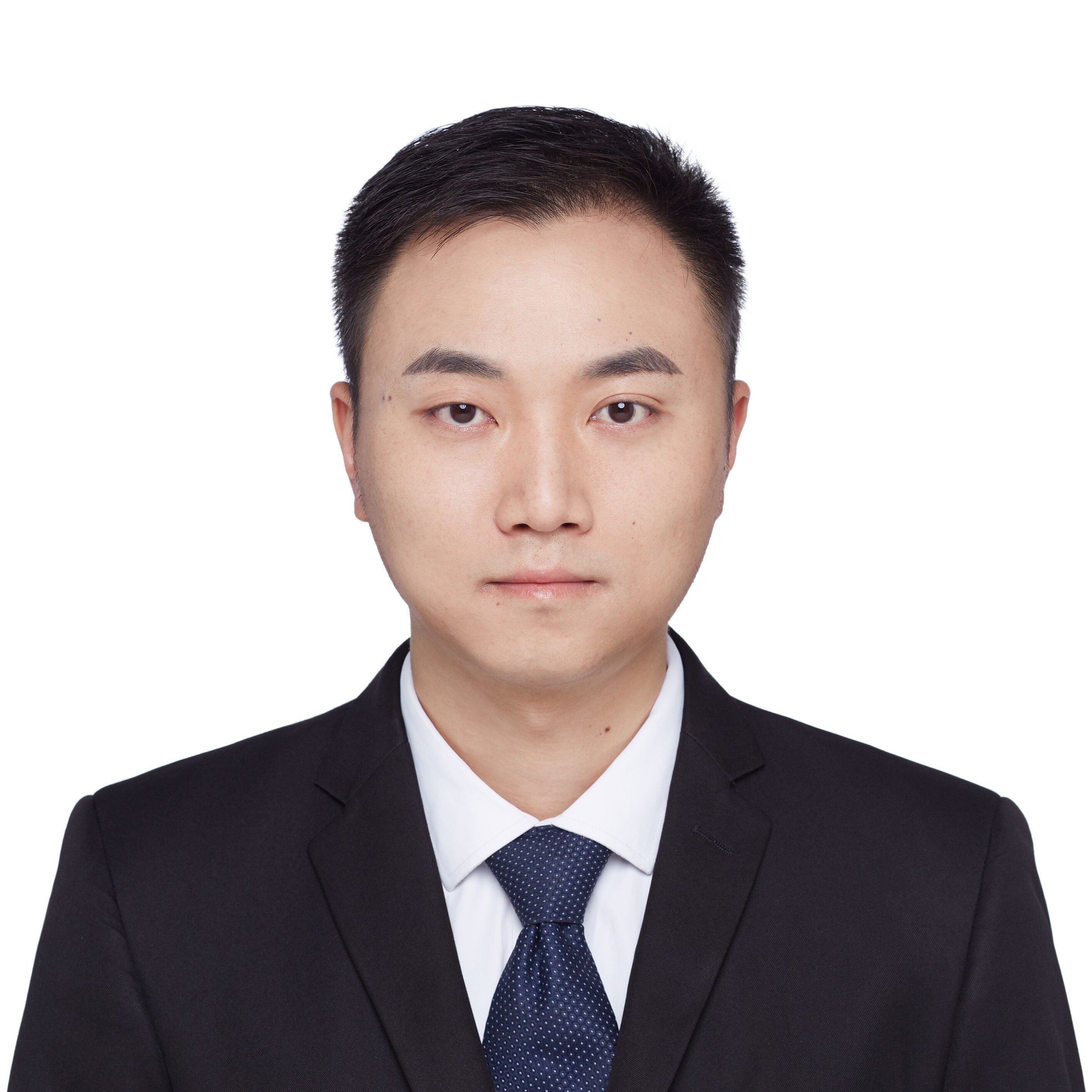}}]{Yan-Pei Cao}
received bachelor's and Ph.D. degrees in computer science from Tsinghua University in 2013 and 2018, respectively. He is currently a principal researcher at ARC Lab, Tencent PCG. His research interests include computer graphics and 3D computer vision.
\end{IEEEbiography}
\vspace{-1em}
\begin{IEEEbiography}[{\includegraphics[width=1in,height=1.25in,clip,keepaspectratio]{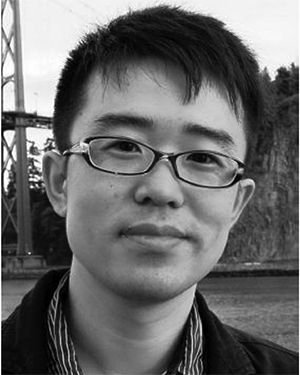}}]{Pengfei Wan}
received a B.E. degree in electronic engineering and information science from the University of Science and Technology of China, Hefei, China, and a Ph.D. degree in electronic and computer engineering from the Hong Kong University of Science and Technology, Hong Kong. His research interests include image/video signal processing, computational photography, and computer vision.
\end{IEEEbiography}
\vspace{-1em}
\begin{IEEEbiography}[{\includegraphics[width=1in,height=1.25in,clip,keepaspectratio]{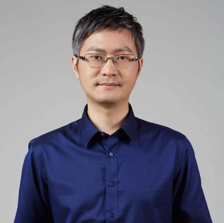}}]{Wen Zheng}
received the bachelor's and master's degrees from Tsinghua University, Beijing, China, and a Ph.D. degree in computer science from Stanford University, in 2014. He is currently the Head of Y-tech at Kuaishou Technology. His research interests include computer vision, augmented reality, machine learning, and computer graphics.
\end{IEEEbiography}
\vspace{-1em}
\begin{IEEEbiography}[{\includegraphics[width=1in,height=1.25in,clip,keepaspectratio]{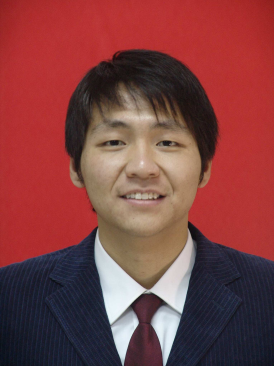}}]{Zhizhong Han}
received his Ph.D. degree from Northwestern Polytechnical University, China in 2017. He was a Post-Doctoral Researcher with the Department of Computer Science, at the University of Maryland, College Park, USA. Currently, he is an Assistant Professor of Computer Science at Wayne State University, USA. His research interests include 3D computer vision, digital geometry processing and artificial intelligence.
\end{IEEEbiography}

\end{document}